\documentclass[final]{cvpr}

\usepackage{times}
\usepackage{epsfig}
\usepackage{graphicx}
\usepackage{amsmath}
\usepackage{amssymb}
\usepackage[export]{adjustbox}

\newcommand*{\ShowNotes}{}

\ifdefined\ShowNotes
  \newcommand{\colornote}[3]{{\color{#1}\bf{#2: #3}\normalfont}}
\else
  \newcommand{\colornote}[3]{}
\fi

\newcommand {\barmayo}[1]{\colornote{red}{BM}{#1}}
\newcommand {\ayellet}[1]{\colornote{blue}{AT}{#1}}

\renewcommand{\[}{\begin{eqnarray}}
\renewcommand{\]}{\end{eqnarray}}
\newcommand{\R}{\mathbb{R}}

\newcommand{\E}{\mathbb{E}}

\usepackage[pagebackref=true,breaklinks=true,colorlinks,bookmarks=false]{hyperref}

\def\cvprPaperID{11061} 
\def\confYear{CVPR 2021}

\begin{document}

\teaser{
\centering
    \begin{tabular}{cccc}  
    \includegraphics[height =3.7cm]{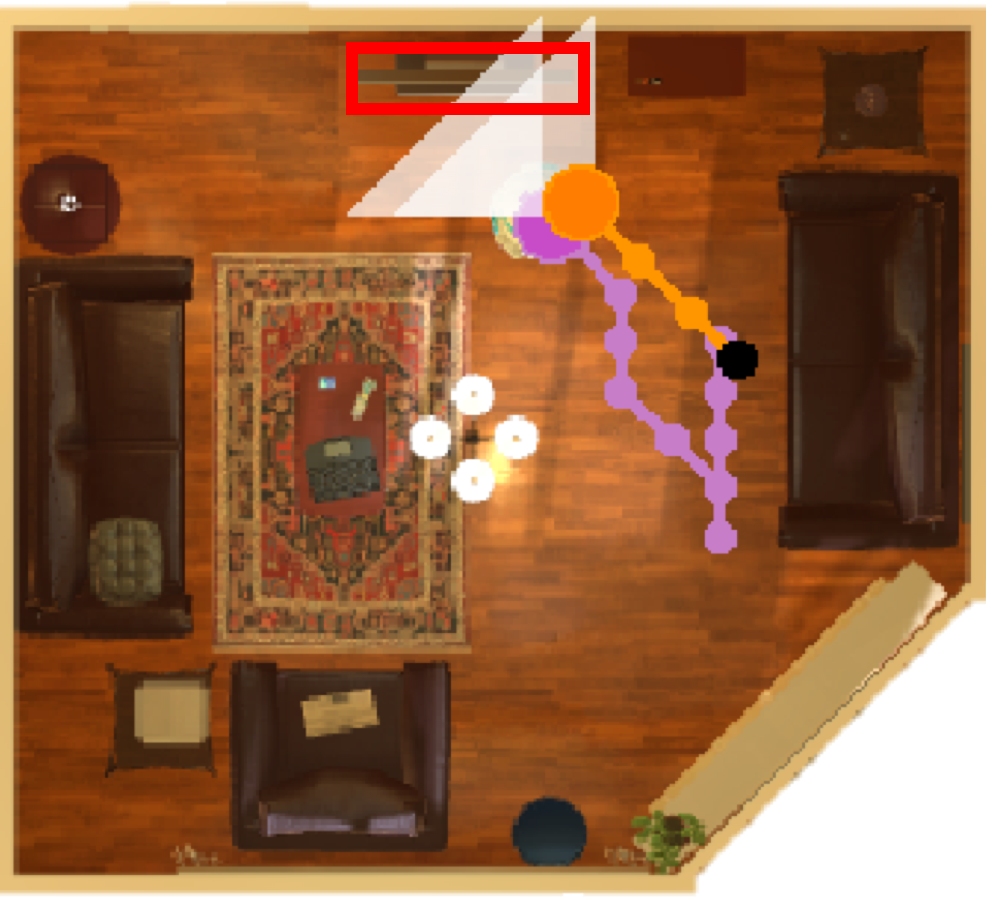} &
    \includegraphics[height =3.7cm]{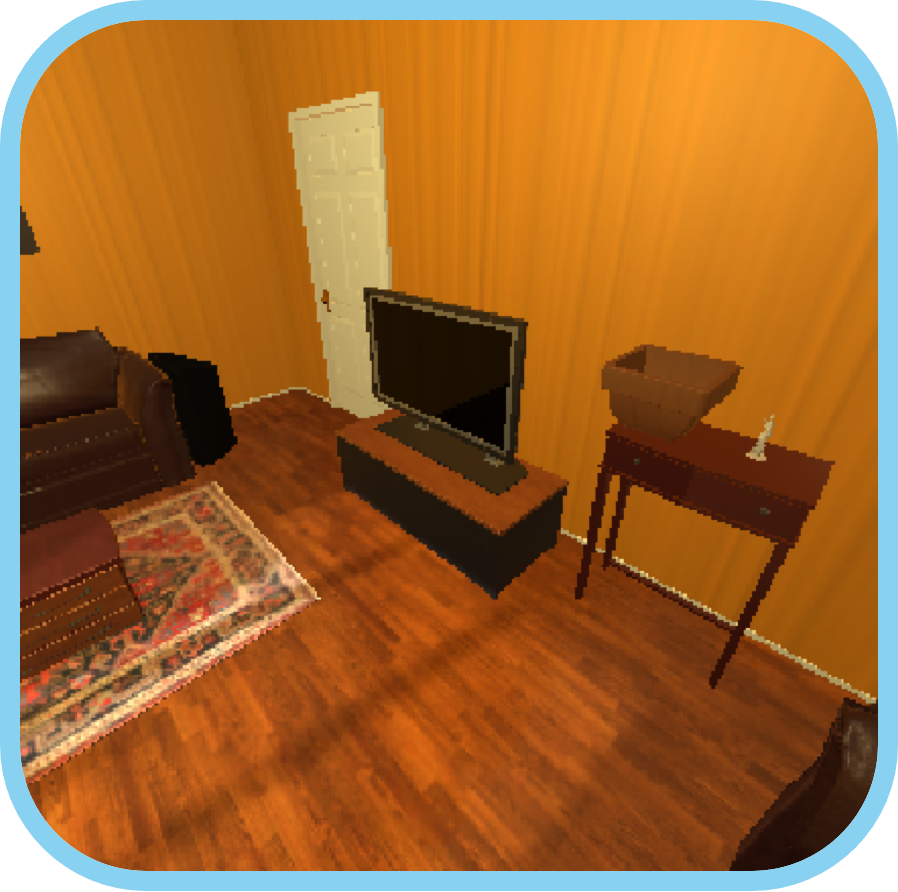} &
    \includegraphics[height =3.7cm]{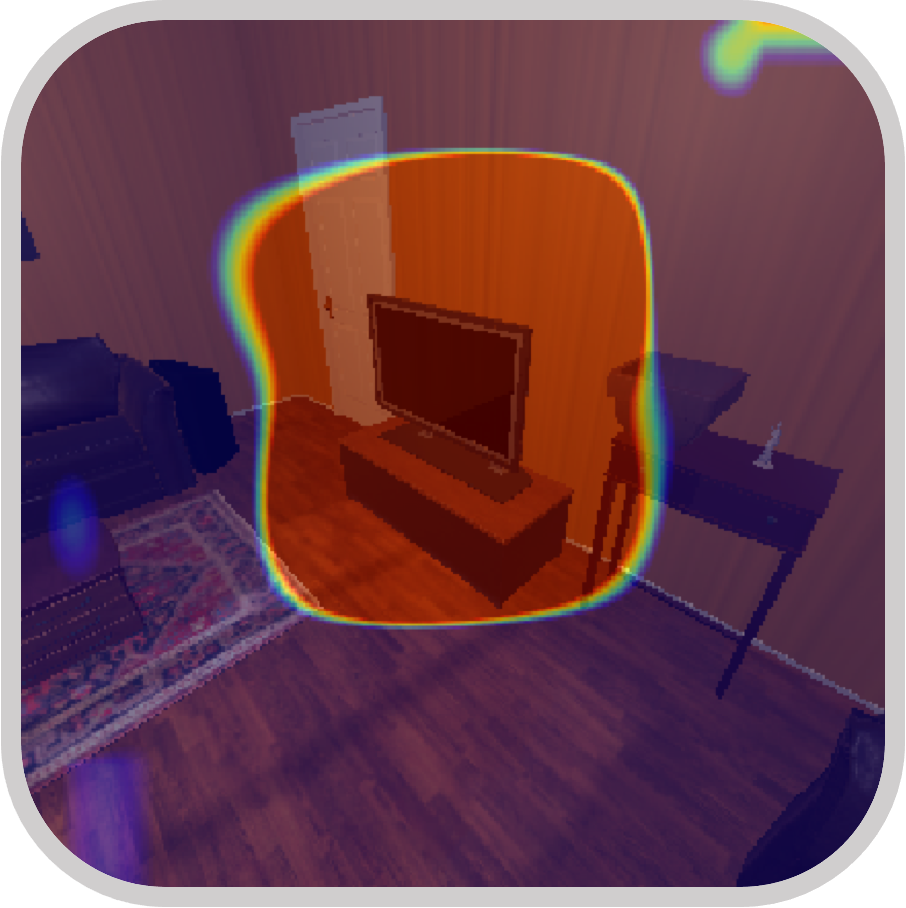} &
    \includegraphics[height =3.7cm]{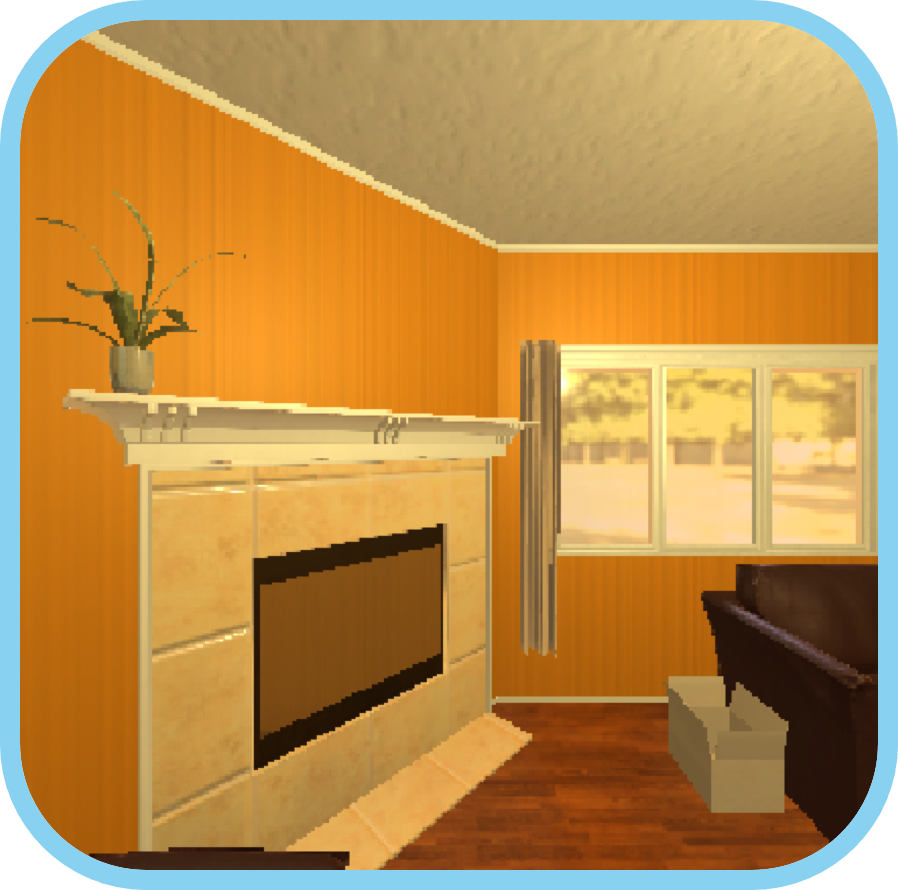} 
    \\
        (a) Paths---Ours \& \cite{Wortsman_2019_CVPR}'s &
    (b) Our agent's view  &
    (c) Our attention &
    (d) \cite{Wortsman_2019_CVPR}'s view\\
  \end{tabular}
\caption{{\bf Visual navigation.} 
(a) The agent aims at finding a TV (red rectangle) in a living room (top view), starting from a given location (black circle).
Our agent's path is marked in orange and~\cite{Wortsman_2019_CVPR}'s path is in magenta. 
At each step, the agent is given a specific view, depending on its position.
In this example, our agent starts by turning around in its starting location to gather information---a strategy it has learned.
(b) shows our agent's view before the first move forward, whereas (d) shows ~\cite{Wortsman_2019_CVPR}'s view before its first move forward.
(c) shows our attention model, which combines semantic and spatial information of (b)'s view;
it directs our agent to move forward, towards the TV.
Differently, the view in (d) is part of \cite{Wortsman_2019_CVPR}'s lengthy exploration (magenta path in (a)) after the sought-after TV. 
}
\label{fig:teaser}
}

\title{Visual Navigation with Spatial Attention}

\author{Bar Mayo\\
Technion\\
{\tt\small mayo.bar@gmail.com}
\and
Tamir Hazan\\
Technion\\
{\tt\small tamir.hazan@technion.ac.il}
\and
Ayellet Tal\\
Technion\\
{\tt\small ayellet@ee.technion.ac.il}
}

\maketitle

\begin{abstract}
This work focuses on object goal visual navigation, aiming at finding the location of an object from a given class, where in each step the agent is provided with an egocentric RGB image of the scene.
We propose to learn the agent's policy using a reinforcement learning algorithm. Our key contribution is a novel attention probability model for visual navigation tasks. This attention encodes semantic information about observed objects, as well as   spatial information about their place. This combination of the ``what" and the ``where" allows the agent to navigate toward the sought-after object effectively. The attention model is shown to improve the agent's policy and to achieve state-of-the-art results on commonly-used datasets. 
\end{abstract}

\section{Introduction}

Human and animals can navigate new environments relatively well. This adaption to new surroundings, although natural, is not trivial. It requires to find parallels between the new observations and our past experience. This is largely possible due to our ability to sort through new visual information and intelligently focus on the most relevant semantic cues. 
For instance, when looking for a toaster in a previously-unvisited kitchen, our intuition is to look for the refrigerator, while ignoring other "irrelevant" information, since our past experience indicates that the toaster is usually located not far from the refrigerator. 

Object goal visual navigation tasks include two basic components: semantic understanding of the scene and path planning \cite{thrun1998learning, blosch2010vision, hadsell2009learning}. With the increase of data and computation power, reinforcement learning algorithms excelled in learning policies for these two components jointly in an end-to-end manner \cite{peters2008reinforcement, mnih2015human, gupta2017cognitive, chaplot2020object}. As a result, many extensions to visual navigation were presented, including tasks specified by natural language instructions \cite{chaplot2017gated}, by a desired goal image \cite{zhu2017icra}, or by a target object \cite{Wortsman_2019_CVPR, du2020learning}. Reinforcement learning of spatial and semantic relations is a fundamental challenge for these tasks \cite{wu2019bayesian}.  

\begin{figure*}[t]
  \centering
  \large
    \begin{tabular}{cccccc}
    \includegraphics[width=0.14\textwidth]{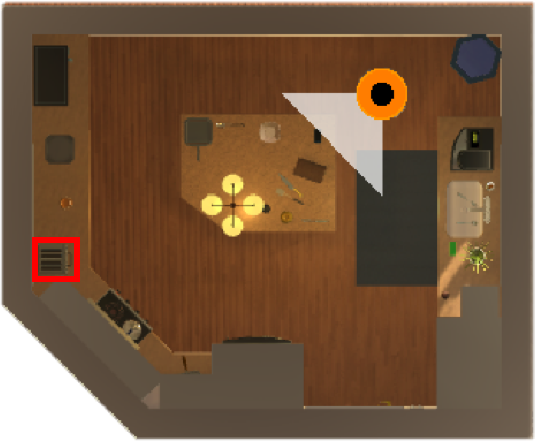} &
    \includegraphics[width=0.14\textwidth]{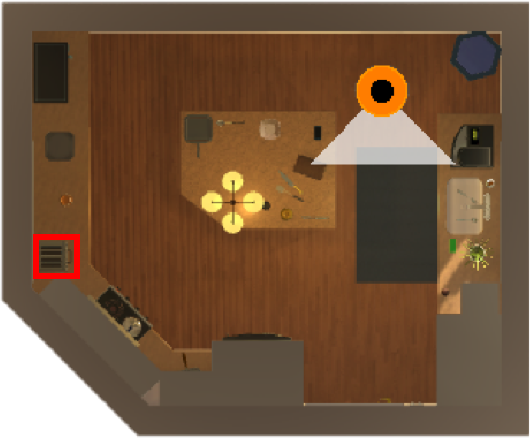} &
    \includegraphics[width=0.14\textwidth]{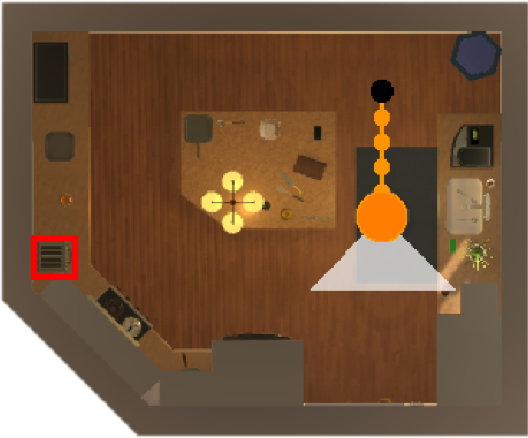} & \includegraphics[width=0.14\textwidth]{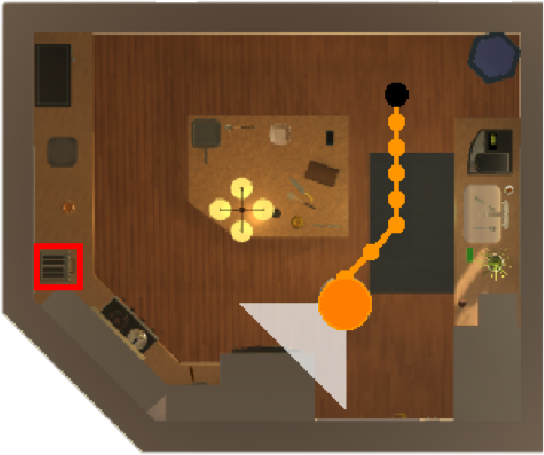} &
    \includegraphics[width=0.14\textwidth]{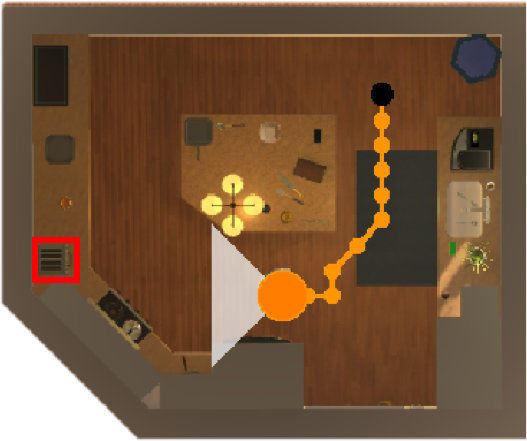} &
    \includegraphics[width=0.14\textwidth]{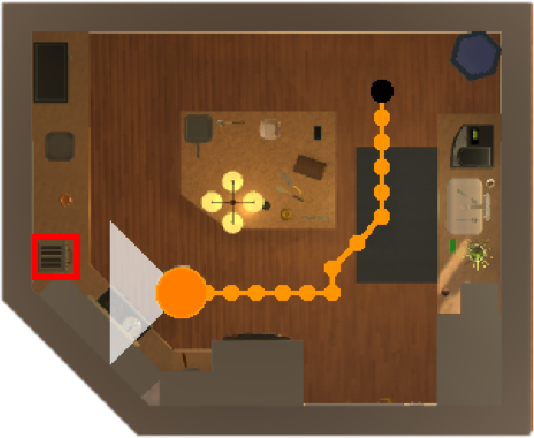} \\
    \includegraphics[height=2.0cm]{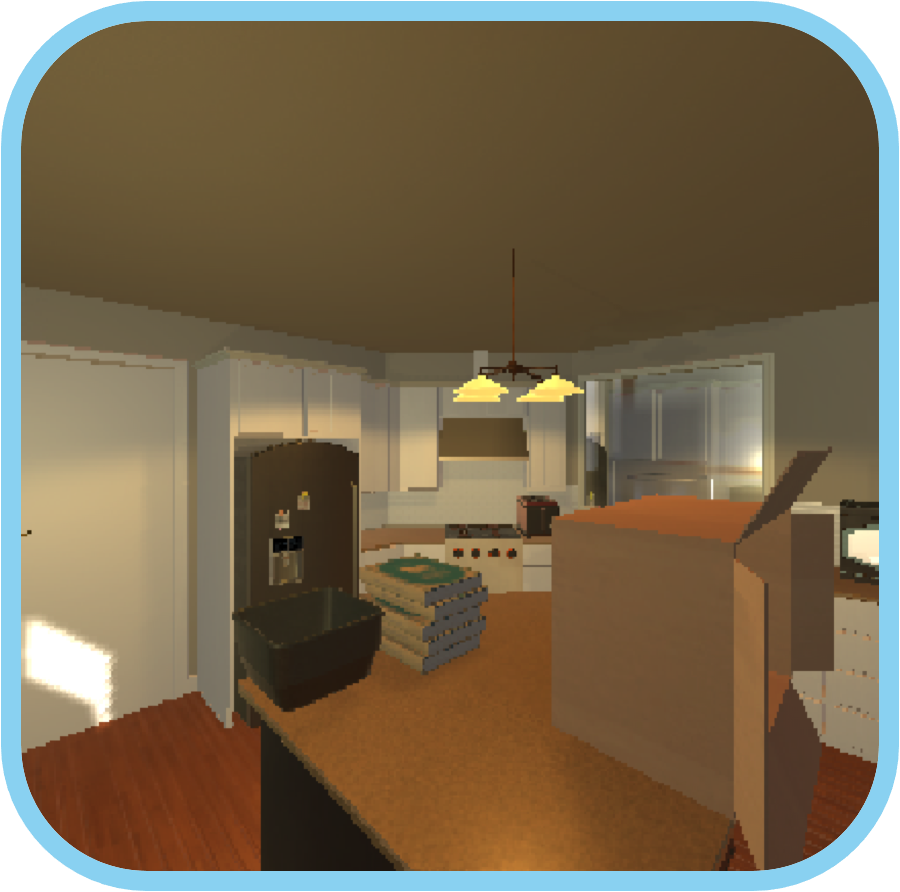} &
    \includegraphics[height=2.0cm]{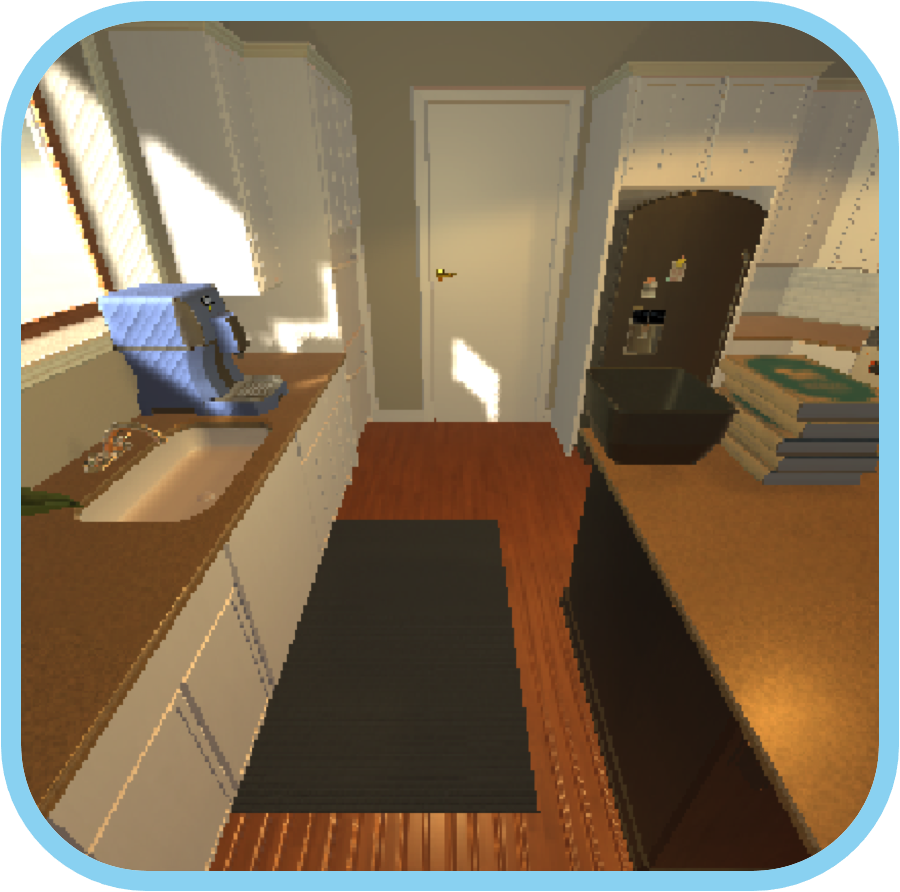} &
    \includegraphics[height=2.0cm]{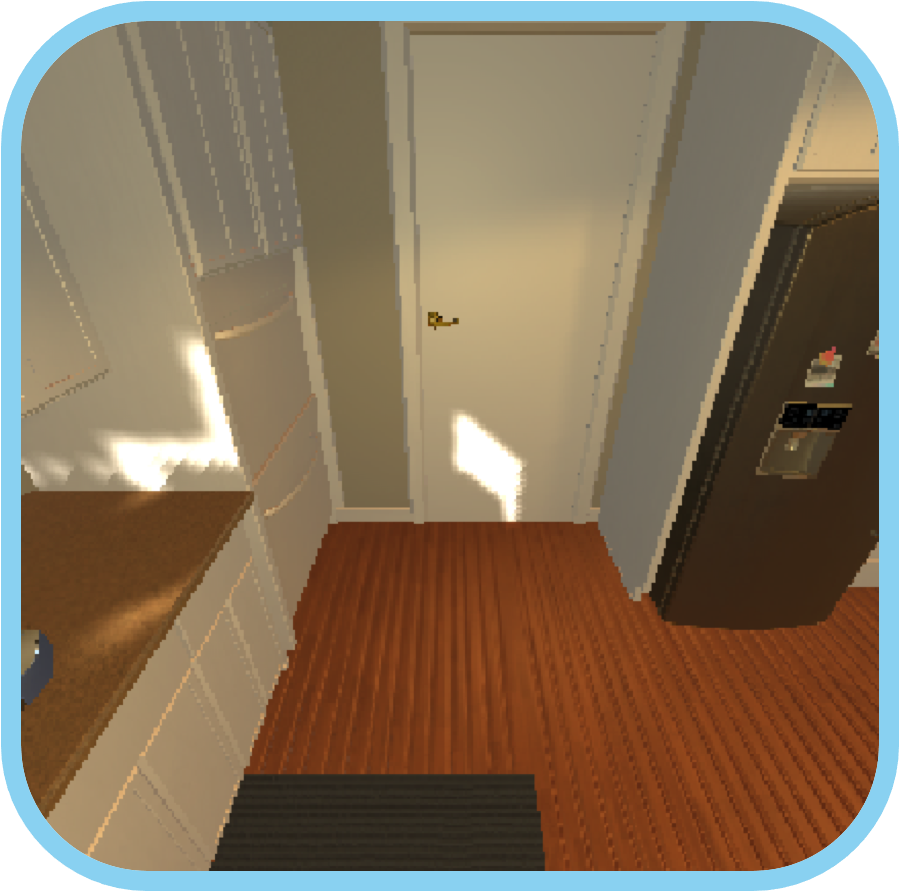} & \includegraphics[height=2.0cm]{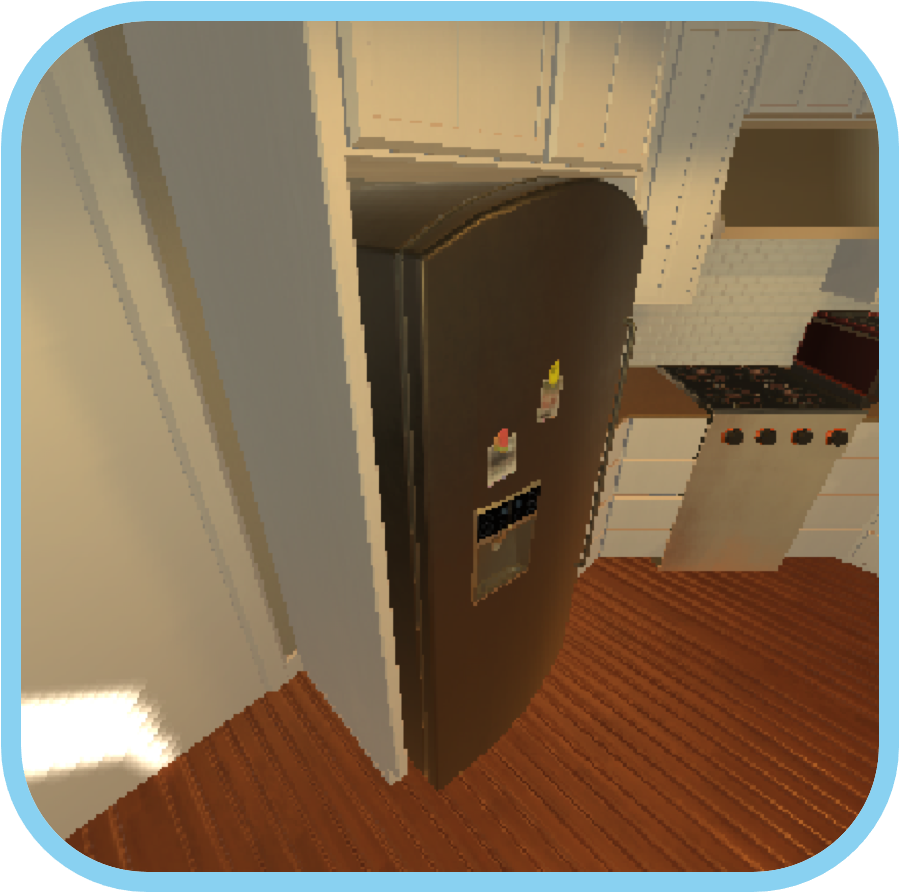} &
    \includegraphics[height=2.0cm]{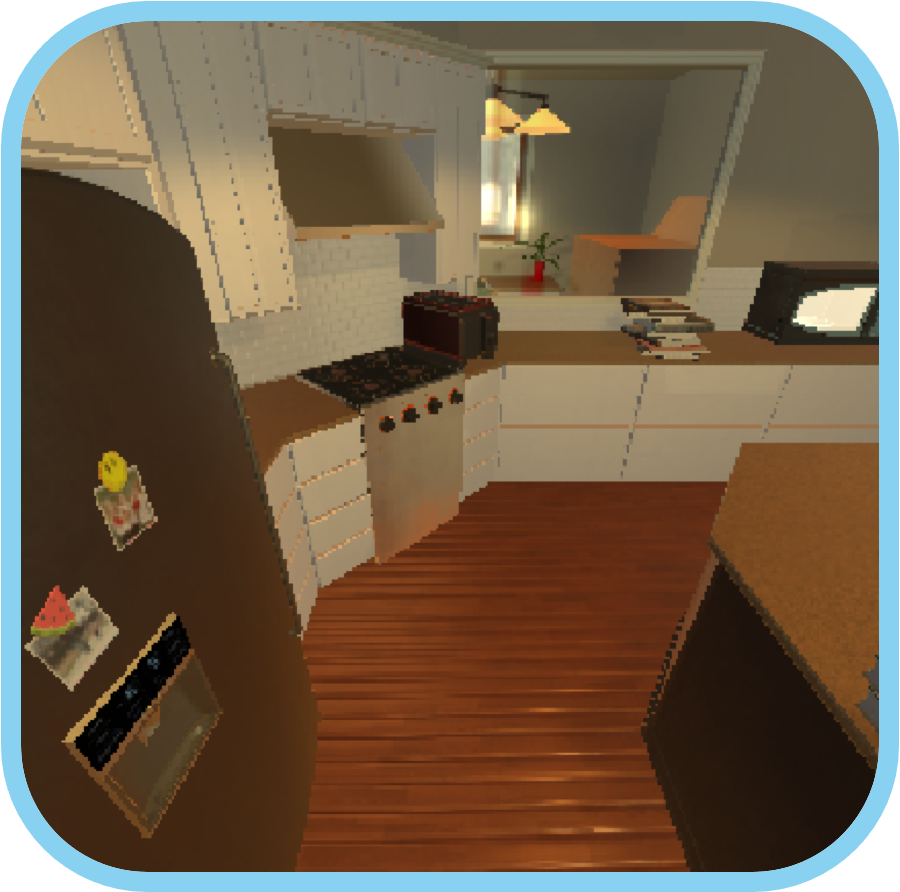} &
    \includegraphics[height=2.0cm]{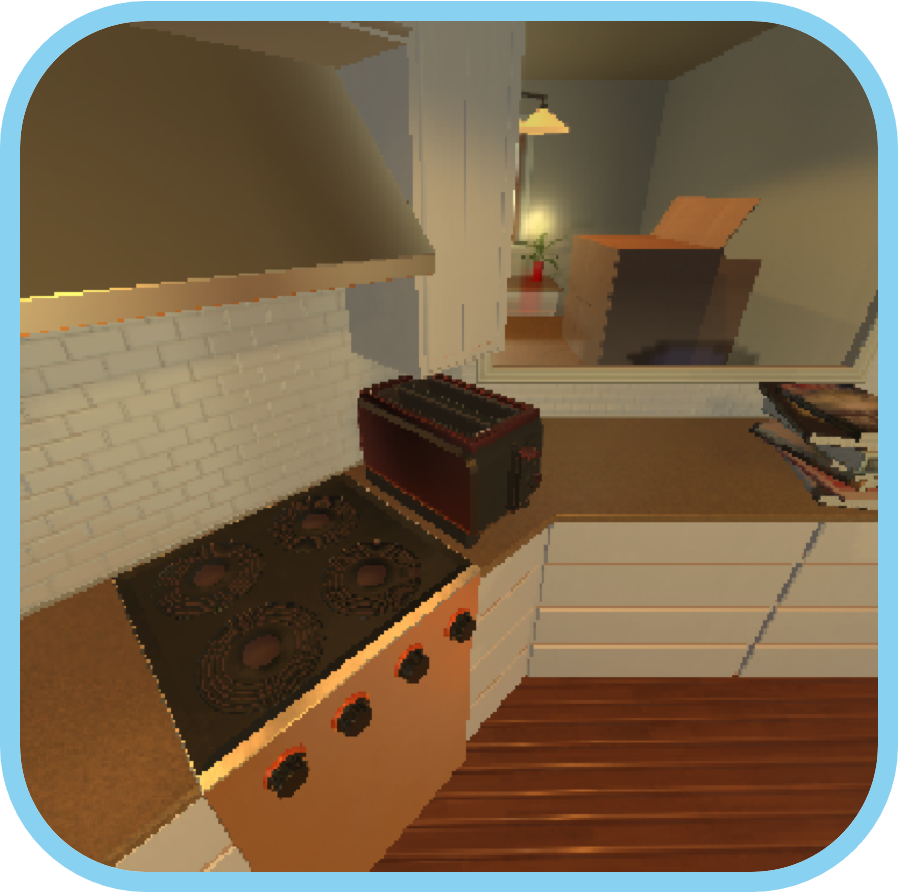} \\
    \includegraphics[height=1.8cm]{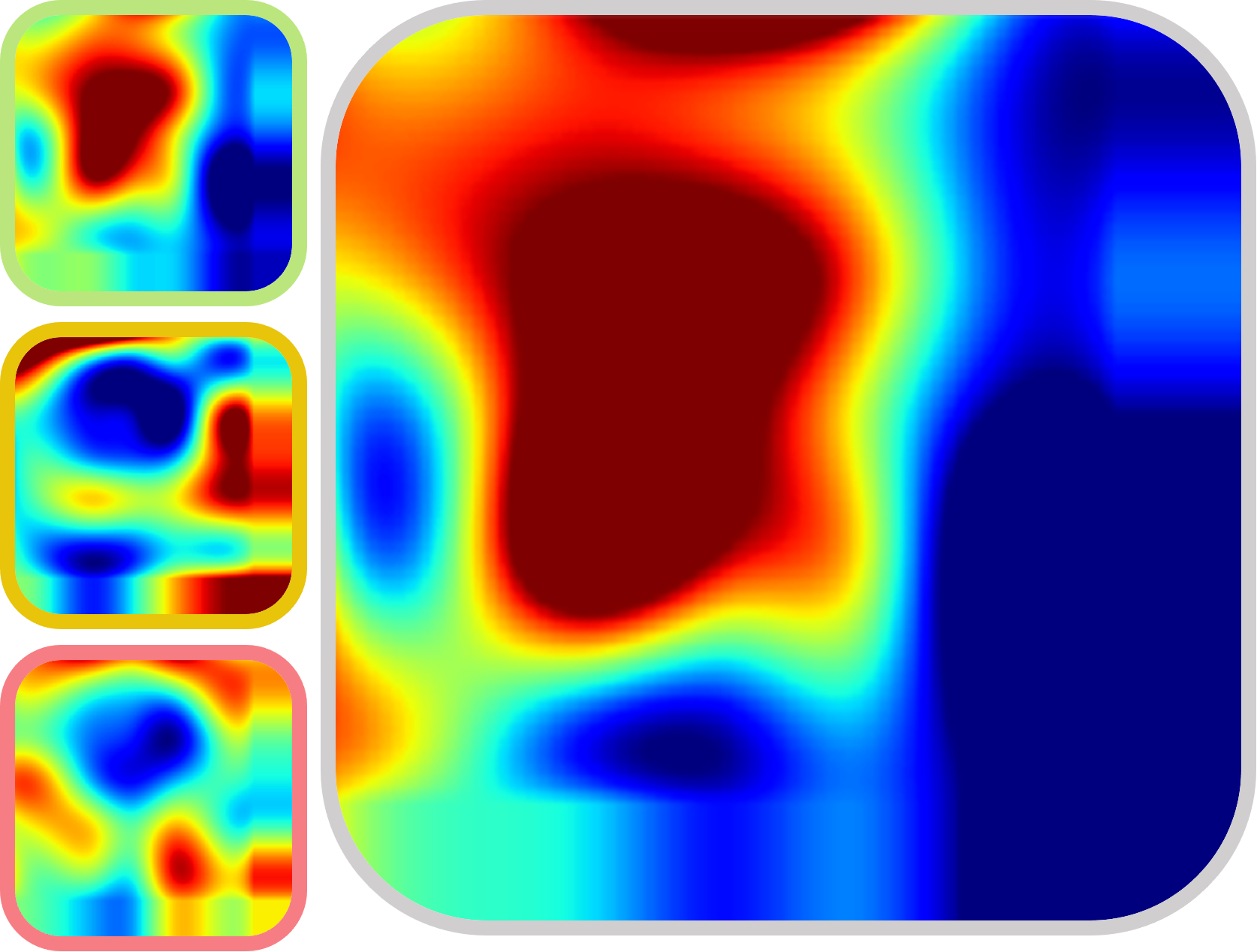}&
    \includegraphics[height=1.8cm]{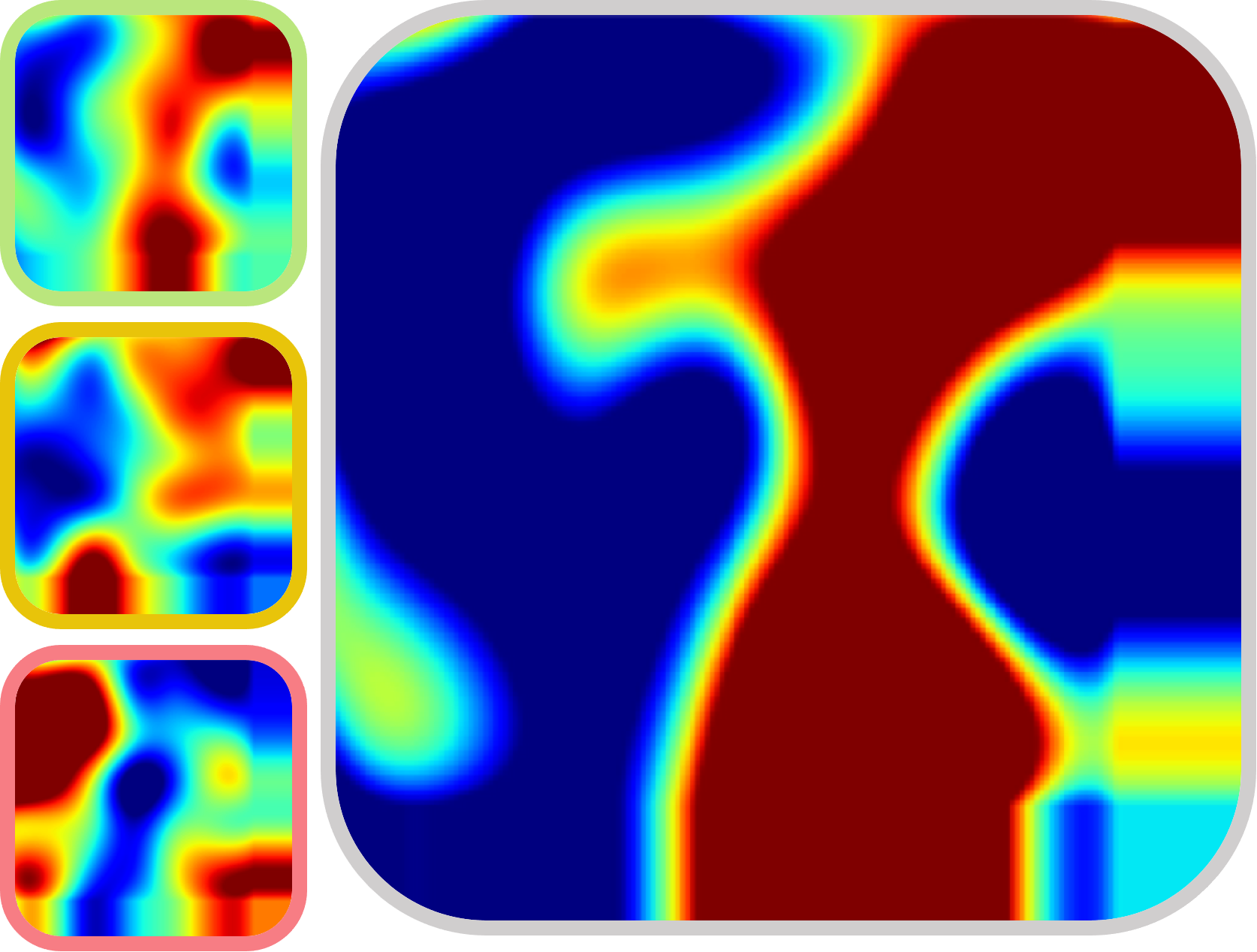}&
    \includegraphics[height=1.8cm]{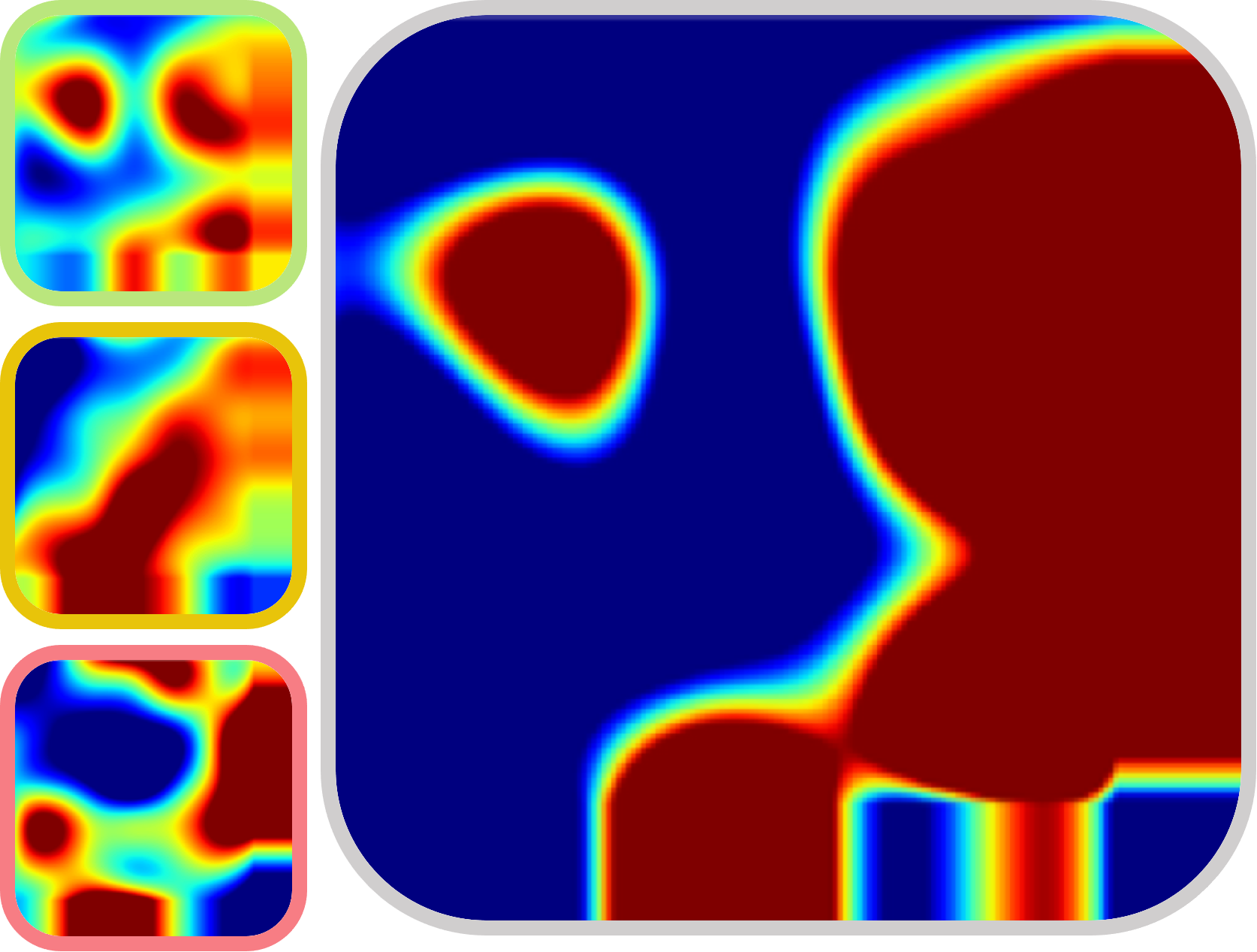}& \includegraphics[height=1.8cm]{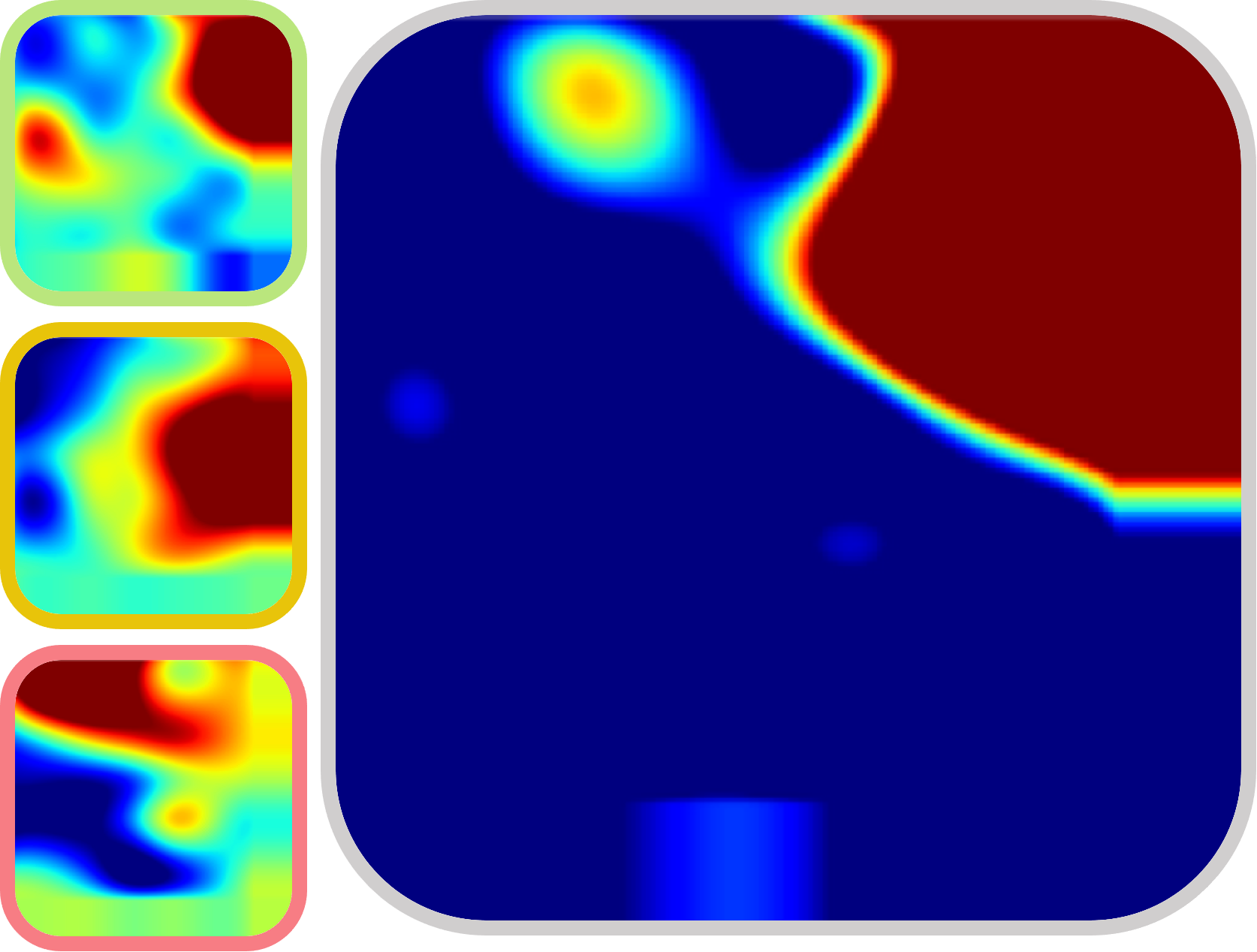}&
    \includegraphics[height=1.8cm]{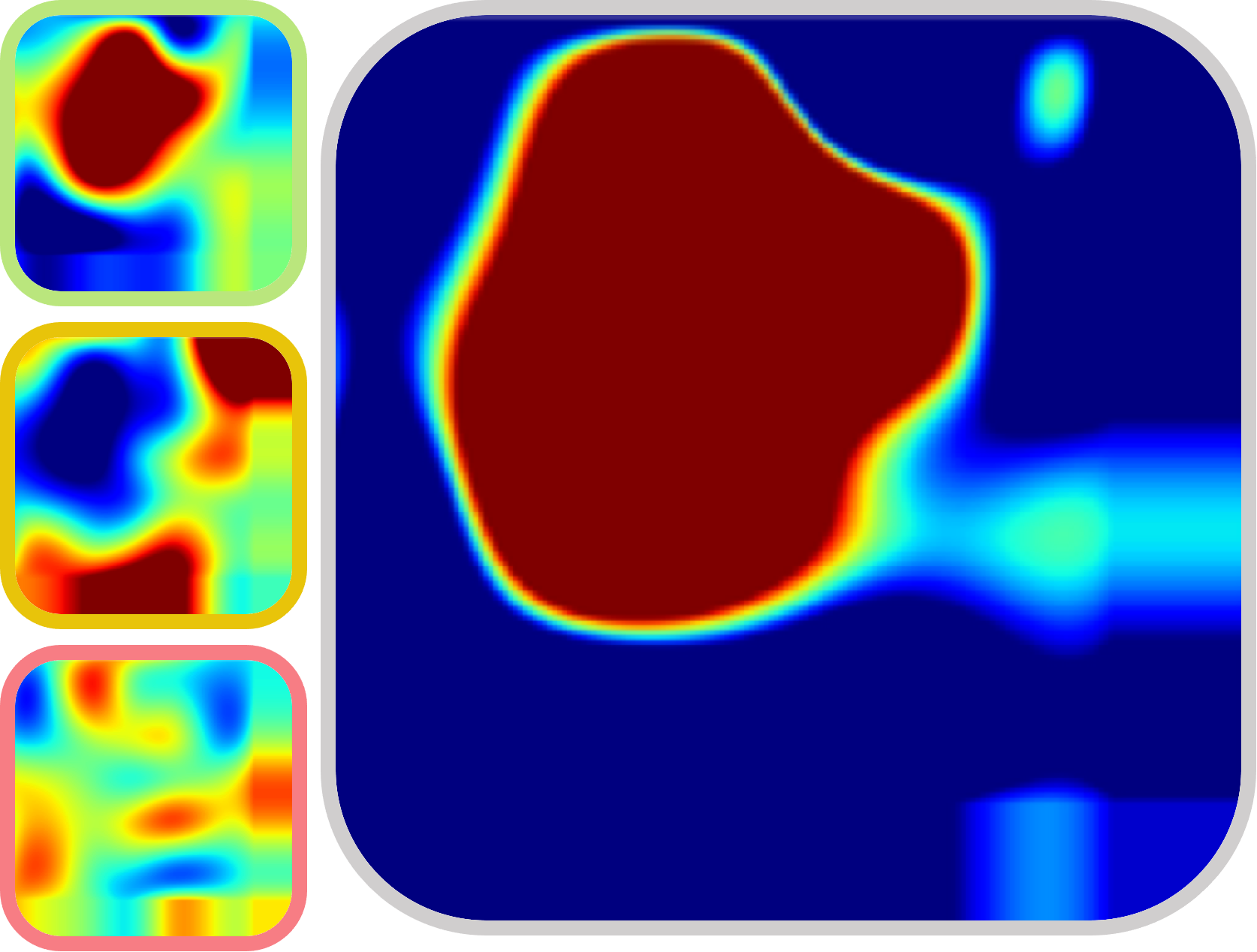} &
    \includegraphics[height=1.8cm]{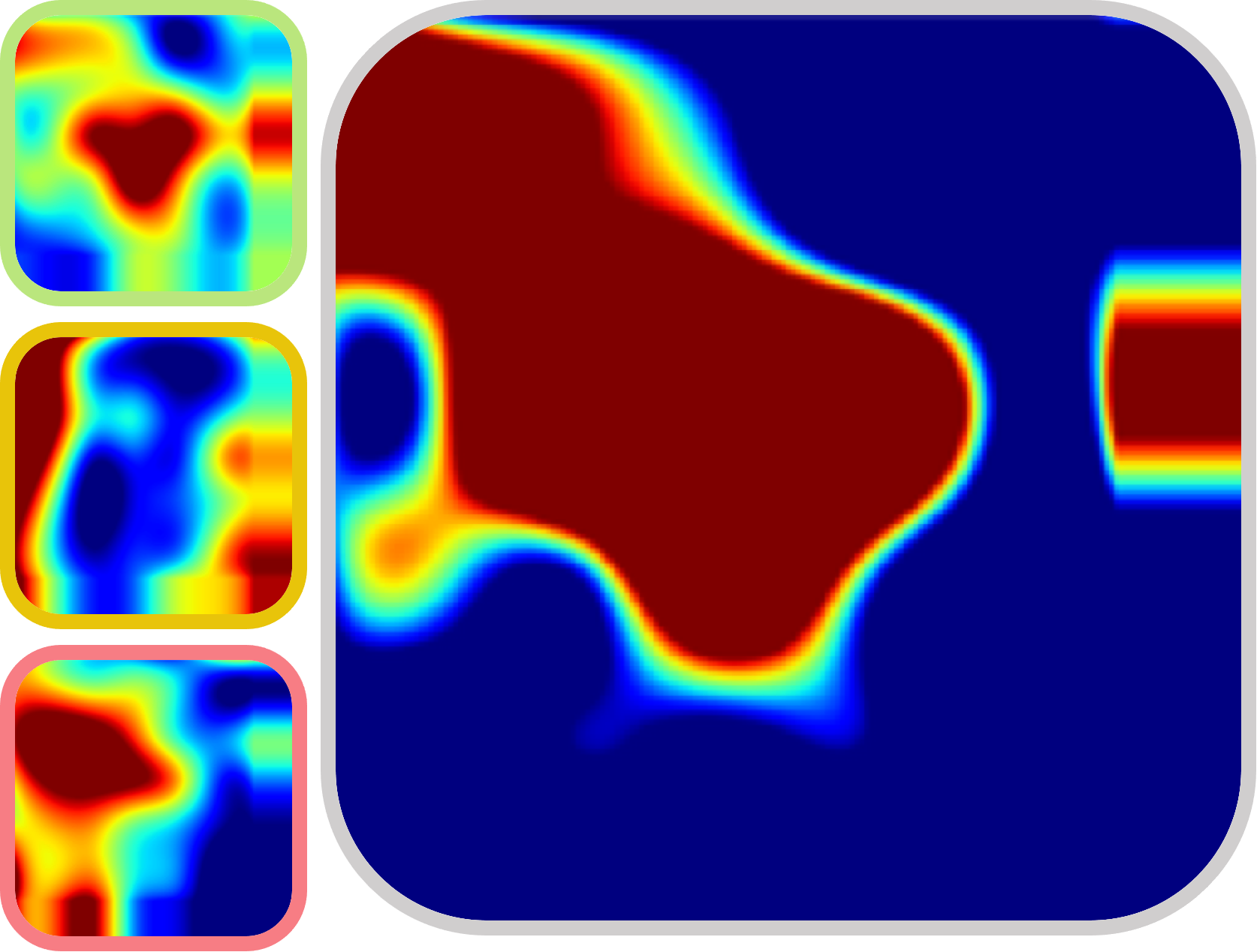} \\
\small{Step $1$} &\small{Step $11$} &\small{Step $16$} & \small{Step $22$} &\small{Step $24$} &\small{Step $29$}
 \end{tabular}
  \caption{{\bf Path sampling.}
  The first row shows a top view of the scene, with the path thus far, along with the agent's view (white triangle) at this step.
  The second row shows the image the agent views at this step.
  The third row shows the fused attention map per step, as well as the three maps that build it.
  The agent is looking for a toaster (red rectangle) and starts from the opposite side of the kitchen.
In Step~$1$ the agent focuses on the refrigerator, which is an indicator to a nearby toaster;
  in Step~$11$ it moves toward the refrigerator;
  in Step $16$ it decides to turns right and then 
  in Step $22$ the toaster becomes visible, at which point the agent's focus switches from the  refrigerator to the toaster and the agent turns right;
  in Step $24$ it starts moving forward, toward the toaster;
  in Step $29$ it is sufficiently close and declares \texttt{Done}.
    }
  \label{fig:toaster_traj}
\end{figure*}

This work focuses on object goal visual navigation, where the goal is to find an instance of the target object class (Figure~\ref{fig:teaser}).
Like previous works, we utilize reinforcement learning.
We propose to improve the agent's policy by encoding semantic information about observed objects using a convolutional net, as well as spatial information about their place, using an attention probability model. This combination of both semantic and spatial information, i.e., of ``what" and of ``where", allows the agent to navigate towards the sought objects effectively. 
Our novel attention mechanism consists of three types of attention probability models for navigation: target attention that considers the target information in the image; action attention that takes into account the last action of the agent; memory attention that considers the agent's previous steps in the scene. Our attention probability model results in an attended embedding, which preserves the semantic and spatial information of objects.


We validate our approach using the AI2-THOR~\cite{zhu2017icra} environment. We use Wortsman et al. \cite{Wortsman_2019_CVPR} setup  with their scenes from four room categories: kitchen, living room, bedroom and bathroom, where an agent is navigating to a given object using only visual observations. In our experimental validation we show that not only we outperform the state-of-the-art, but also our attention unit carries spatial information about the objects. This is achieved using a probability distribution over areas of the observed image that are represented by the spatial locations of the topmost convolutional neurons of a standard convolutional net (e.g., ResNet18). As this attention probability distribution preserves the spatial information that is fed to the reinforcement learner, it controls the areas of the image that the agent considers when improving its policy. Hence, this attention unit also carries the promise to explain the agent's actions in visual navigation tasks. 

Figure~\ref{fig:toaster_traj} illustrates this promise.
For instance, in Step~$1$  the attention map suggests that the agent focuses on the refrigerator, which is a good indicator to the location of the toaster.
Similarly, once the toaster becomes visible in Step~$22$, the attention map switches from focusing on the refrigerator to focusing on the sought-after toaster, and in accordance with that, the agent turns right.

Hence, this paper makes three contributions:
\begin{enumerate}
\item
We propose a novel attention mechanism that suits navigation.
It consists of three types of attentions: target, action, and memory.
\item
We present an end-to-end reinforcement learning framework that realizes the attention mechanism and achieves state-of-the-art results.
\item
An added benefit of the different attention maps is being able to explain the agent's actions through visualization. 
\end{enumerate}


\section{Related Work}
\label{sec:related_work}

Navigation is one of the most fundamental problems in mobile robotics. Traditional navigation approaches decompose the problem into two separate stages: mapping the surrounding and planning a path to the goal \cite{blosch2010vision, cummins2007probabilistic, dissanayake2001solution, hadsell2009learning,  kidono2002autonomous, thrun1998learning}. Generally, these works treat navigation as a purely geometric problem.  Reinforcement learning (RL) methods were applied to learn policies for robotic tasks \cite{kim2004autonomous, kohl2004policy, mnih2015human, peters2008reinforcement}. While RL methods are able to learn complex tasks in an end-to-end manner, their main challenge in visual navigation tasks is to understand both the visual cues as well as the navigation plan. 
Recently, Shen et al.~\cite{shen2019situational} fused different visual representations in navigation. In a related thread, Gupta et al.~\cite{gupta2017cognitive} developed a cognitive mapping and planning approach whose map. Similarly, \cite{georgakis2019simultaneous, cartillier2020semantic} build semantic maps in a pre-exploration setting to capture spatial information in visual navigation. 
While our approach also uses a latent spatial representation, it differs in important respects: our spatial information relies on an attention probability distribution over areas in the image. This component serves as an important building block in our attended embedding, which combines both the spatial and semantic information of the image.  

Target-driven visual navigation tasks have been proposed to search for an object in visual scenes. Zhu et al. \cite{zhu2017icra} address target-driven navigation given a picture of the target, while Mousavian et al. \cite{mousavian2019visual} augment the learner with semantic segmentation and detection masks. Chaplot et al. explores visual navigation given language instructions using gated attention \cite{chaplot2017gated} and semantic mapping \cite{chaplot2020object}. In contrast to our work, they use an attention module to represent their language modality, while we use attention probability distribution over areas of the image to better understand the spatial information. More broadly, Bayesian methods for visual navigation with that relation graphs appear in \cite{wu2019bayesian, anderson2019chasing}. 

We validate our visual navigation approach on AI2-THOR~\cite{zhu2017icra}, which is an environment that consists of near photo-realistic $3$D indoor scenes \cite{kolve2017ai2}. We augment the work on self-adaptive visual navigation (SAVN) of Wortsman et al., \cite{Wortsman_2019_CVPR} with attended observation that serves an input to its model agnostic meta-learner (MAML) \cite{finn2017model}. Other approaches for visual navigation were applied to AI2-THOR, e.g., learning scene priors using graph convolutional nets that are able capture the relationships between objects in the scene \cite{yang2018visual}. Recently, Du et al. augmented the AI2-THOR environment with detection information \cite{du2020learning} albeit for different scenes. 

Our work develops an embedded attention module that combines both semantic and spatial information \cite{vaswani2017attention, schwartz2019factor}. The spatial information is encoded by an attention probability distribution over areas in the image and the semantic information of these areas is encoded by a convolutional net. Attention in visual tasks has mainly been deployed for language augmented tasks \cite{bahdanau2014neural,jetley2018learn, mun2016text, xu2016ask, xu2015show,  yang2016stacked, you2016image, blukis2018mapping, lee2020perceptual}. Similar to our work, they construct an attention probability distribution over areas of the image. However, these attention units typically summarize the convolutional net representation by averaging with respect to the attention probability distribution. In contrast, we refrain from averaging and preserve the spatial dimension of the convolutional layer, which significantly improves its performance on navigation tasks. Similar to our work, natural language attention models, and in particular multi-head attention \cite{vaswani2017attention}, use probability models to re-embed their preceding layer. However, they do not retain the spatial information of the image and their attended embedding ignores this information.

\section{Attended Navigation}
\label{sec:model}

Our navigation task $ \tau \in {\cal T}$ considers a scene $S$, a starting point $p$ and a target object $o$. The goal of a task $\tau = (S,p,o)$ is to move an agent in a 3D indoor scene from the starting position $p$ to an instance of the target object class $o$ with a minimum number of steps.

The navigation is preformed by a mobile agent and is learned by a policy. The agent's policy is limited to six actions: \texttt{MoveAhead}, \texttt{RotateLeft}/\texttt{RotateRight}, \texttt{LookDown}/\texttt{LookUp}, \texttt{Done}.  At each step the agent is given an egocentric RGB image $s \in S$ from the scene $S$ and a target object class $o$ and it can act in one of two ways:
(1)~choose one of the possible movement directions and move accordingly or
(2)~issue the \texttt{Done} action, signalling that the agent “believes” it has finished the task. The \texttt{Done} action ends the trial of navigation, termed an {\it episode}.

An episode is finished successfully if the agent issued a \texttt{Done} action and
(1)~The target object is sufficiently close to the agent ($1$ meter in practice);
(2)~The target object is in view; and
(3)~The agent did not pass the maximum number of allowed steps.

Following the {\it Self-Adaptive Visual Navigation (SAVN)} framework of Wortsman et al. \cite{Wortsman_2019_CVPR}, we learn a policy $\pi_\theta(\cdot | s)$, which chooses an action $a$ given an egocentric RGB image $s$ within a scene~$S$. We use gradient decent (policy gradient) to improve the policy's parameters $\theta$ to navigate in each episode. These parameters are learned in order to maximize the expected reward $\E [{\cal R}^\tau]$ on a sequence of actions in a given episode. In our experimental evaluation, we use the SAVN navigation reward  ${\cal R}^\tau_{\text{nav}}$ that subtracts $0.01$ for any step except \texttt{Done} and adds $5$ for a successful navigation. We also use the actor-critic family algorithms that minimize its navigation loss ${\cal L}^\tau_{\text{nav}}(\theta,a)$, which consists of the negative expected reward that serves the actor and a learned value function that serves the critic.

\begin{figure*}[t]
\centering
\includegraphics[width=1\textwidth]{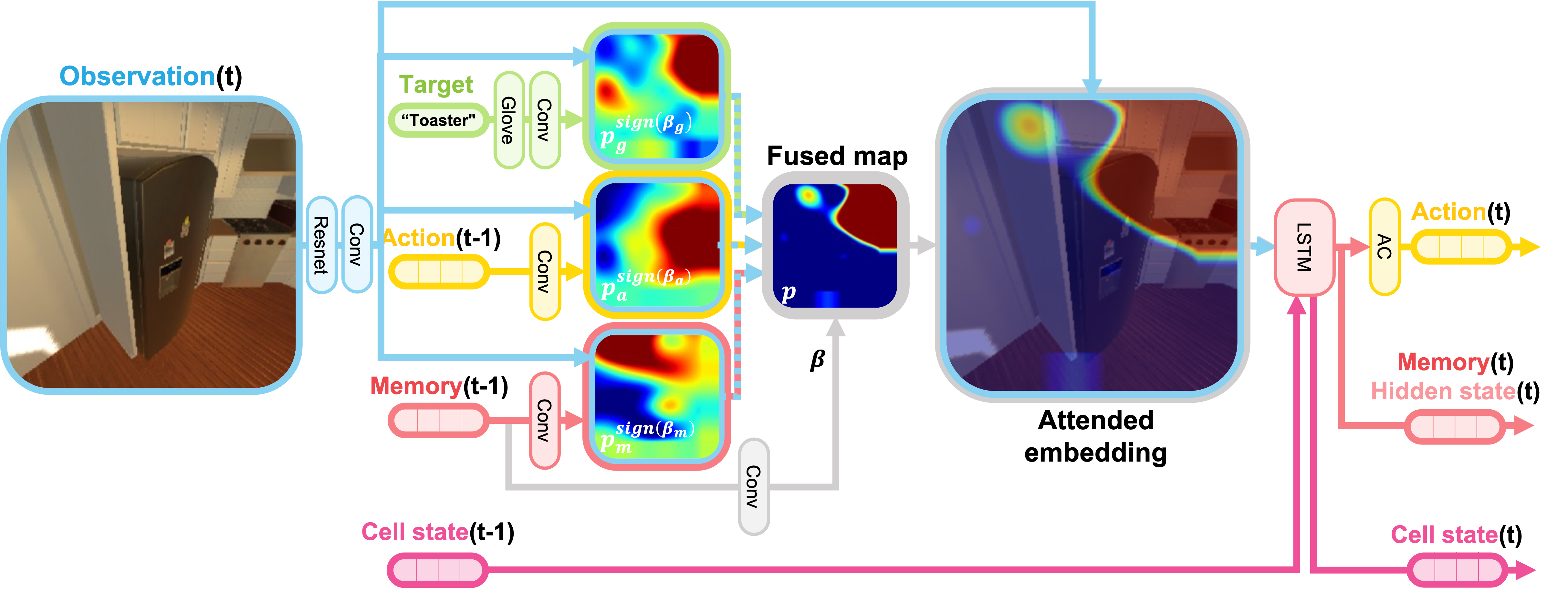}
\caption{ 
{\bf Architecture overview.} The adaptive navigation unit, which is described in Section \ref{sec:adaptive}  follows Wortsman et al. \cite{Wortsman_2019_CVPR}. 
The attended embedding, described in Equation \ref{eq:embedding},  encodes semantic information about observed objects using a convolutional net, as well as spatial information about their place, using the fused attention probability distribution. 
The fused attention, described in Section \ref{sec:fused} balances the target/action/memory attention distributions. 
The target attention, described in Section \ref{sec:target}, combines the target word GloVe embedding with the image information. 
In this example, the target word is ``toaster" and  $t=22$.
One can verify that the inferred  probability distribution overlaps the area in the image that contains part of the toaster (the black rectangle to the right). The action attention, described in Section \ref{sec:action}, combines the last action of the actor with the image information.  In this example, the action's attention probability distribution is focused on the area on the right of the image, and the agent is about to turn right to locate the toaster. The memory attention, described in Section~\ref{sec:memory}, summarizes the agent's experience and aims to focus on sections of the image based on the information already gathered in the episode. In this example, the memory attention probability distribution takes into account the refrigerator, as it was  learned to be a correlated to the toaster class in a kitchen. 
}
\label{fig:architecture}
\end{figure*}

\subsection{Adaptive navigation}
\label{sec:adaptive} 
SAVN~\cite{Wortsman_2019_CVPR} relies on adaptive navigation, hence its policy benefits from adapting to the relevant navigation subtask, e.g., entering a hallway, approaching a refrigerator and so on. To deal with such a complex task, SAVN applies model agnostic meta-learning (MAML) that shifts the parameters of the policy as the agent interacts with the scene. This shift of parameters allows the agent to adapt to the scene while interacting with it. SAVN achieves this behavior by using an interaction loss ${\cal L}^\tau_{\text{int}}(\theta,\alpha)$ that is being applied on a $\hat k$-prefix $\alpha$ of actions $a$, i.e., $\alpha = (a_1,...,a_{\hat k})$. Thus the loss function for learning a policy $\pi_\theta(\cdot | s)$ for a task $\tau$ for a sequence of actions $a$ and their prefix $\alpha$ is 
\[
\min_\theta \sum_{\tau \in {\cal T}_{\text{train}}} \E_{a \sim \pi_\theta} \Big[ {\cal L}^\tau_{\text{nav}}  \big( \theta - \nabla_\theta {\cal L}^\tau_{\text{int}}(\theta,\alpha) ,a \big) \Big].
\]
We also learn parameters $\phi$ of the interaction loss, although we omit this dependence for readability. 
This loss essentially minimizes the navigation loss while encouraging the gradient $\nabla_\theta {\cal L}^\tau_{\text{int}}(\theta,\alpha)$ to be similar to the gradient $\nabla_\theta {\cal L}^\tau_{\text{nav}}(\theta,a)$.  This allows to adjust the policy parameters in test time, to reduce the navigation loss of a single trajectory, i.e., to better adapt the policy to navigation subtasks.   

We introduce spatial attention into the SAVN framework.
Intuitively, attention may improve navigation by orienting  the agent to the correct direction.
We show that this is indeed the case and that we outperform SAVN using a spatial attention mechanism that takes into account the target, the agent's actions, and the memory of images seen so far.
Hereafter, we present our novel attention mechanism, designed particularly for efficient visual navigation in $3$D, and explain how we incorporated it into the architecture. 


\subsection{Spatial embedding using attention}

Visual navigation requires not only semantic reasoning, but also spatial reasoning. This is due to the fact that we control an agent that interacts with a $3$D environment. 
In our work we learn a policy for navigating in the $3$D space given an egocentric RGB image. 
In the following we present an approach that is able to encode semantic information about observed objects using convolutional net, as well as spatial information about their place, using an attention probability model. 
Our approach is illustrated in Figure~\ref{fig:architecture}. 

The navigation is preformed by a mobile agent and is learned by a policy $\pi_\theta(\cdot | s_0)$, which chooses an action given an egocentric RGB image~$s_0$ at the beginning of the episode, within a scene $S$. The policy samples its actions iteratively. At time $t$ the agent is given the egocentric image $s_t$ and chooses the action $a_t$. 

We use convolutional nets to extract semantic information about a given image in the scene, as they were proven to be very effective in encoding mid-level and high-level semantic information in the image. We encode the $t^{th}$ image by the spatial locations of the topmost convolutional layer, whose dimension is $n_v \times n_v \times d_v$, of a standard convolutional net (ResNet18) that is pretrained on Imagenet. The spatial location of each topmost convolutional neuron is indexed by $i,j = 1,...,n_v$ and its $(i,j)^{th}$ location corresponds to an area in the observed image and is described by the vector  $v^t_{i,j} \in \R^{d_v}$. In the following we refer to the area that is represented by such $(i,j)^{th}$ neuron, namely $v_{i,j}^t$, as the $(i,j)^{th}$ sub-window in the image. We then emphasize the spatial information of the objects in the relevant sub-windows using an attention probability distribution. 

At each time step of the agent, we construct an attention probability distribution over the $n_v \times n_v$ sub-windows of the input image. 
Intuitively, this probability distribution assigns high probability to sub-windows that have relevant information in the image and assigns low-probability to sub-windows that do not. By doing so, the attention probability distribution introduces spatial information to the process. Our attention probability distribution is composed of three attention units: (i)  target attention unit, which incorporates the target information in the image; (ii) action attention unit, which takes into account the agent's last action; (iii) memory attention unit, which ``remembers" relevant information from previously-seen images in the scene. These three distributions over the $n_v \times n_v$ sub-windows are then fused into a single attention probability distribution over the image sub-windows. We denote by $p^t(i,j)$ the fused probability distribution at time $t$ over the $n_v \times n_v$ sub-windows $i,j = 1,...,n_v$. 

The spatially attended embedding, $\hat v^t_{i,j}$, of the $t^{th}$ image combines both the semantic information in the image as well as the spatial information about the location of the different objects. The semantic information is represented by the vectors $v^t_{i,j} \in \R^{d_v}$, while the spatial information is represented by the attention probability distribution $p^t(i,j)$. We combine these two components using the pointwise multiplication: $\hat v^t = p^t \odot v^t$, which is defined by 
\[
\hat v^t_{i,j} = p^t(i,j) \cdot v^t_{i,j}.
\label{eq:embedding}
\]
The dimension of the attended embedding is the same dimension as the image embedding $v^t_{i,j}$. Intuitively, the attention probability distribution $p^t(i,j)$ has high values for relevant $(i,j)^{th}$ sub-windows of the image, i.e., sub-windows that contain semantic information for the visual navigation task. Equivalently $p^t(i,j) \approx 0$ for irrelevant sub-windows. Hence, the attended embedding in Equation \ref{eq:embedding} consists of the vector $\hat v^t_{i,j} \approx 0$ whenever $p^t(i,j) \approx 0$, i.e., for sub-windows that are irrelevant for navigation in the $t^{th}$ step. Equivalently, for semantically meaningful sub-windows the attended embedding is similar to the original image embedding, i.e., $\hat v^t_{i,j} \approx v^t_{i,j}$. This embedding highlights the spatial locations of the semantically meaningful sub-windows and populates them with the respective semantic information of the image. This embedding allows the agent to choose its next step according to both the semantic and the spatial information of the image, as it is fed as the input to the navigation policy; see Figure \ref{fig:architecture}.

\subsubsection{Target attention unit} 
\label{sec:target}
This unit learns a probability distribution function over the image. 
It gets as input the image at the $t^{th}$ step and the target  (given by a word) and 
aims to focus on target-relevant information in the image, including the target and visual clues for the target's location. 
For example, if the target is a soap bottle, which is invisible, the agent should focus on the bathtub or the sink, since soap bottles are usually found next to them.
In other words, we want to learn the interaction of each sub-window in the image with the target. 
%

The target word is encoded by a vector of length $d_g$; in our system we used the GloVe embedding~\cite{pennington2014glove}.
We denote by $u_g \in \R^{d_g}$ the GloVe embedding and by $v^t_{i,j} \in \R^{d_v}$ the $n_v \times n_v$ image vectors at the $t^{th}$ time step, for $i,j =1,...,n_v$. The interaction of the word vector $u_g$ with an image sub-window embedding $v^t_{i,j}$ relies on the inner product of these vectors, after embedding both representations in a $d$-dimensional space.

Let $W_v \in \R^{d \times d_v}$ be trainable parameters that embed a sub-window embedding $v^t_{i,j}$ in the $d$-dimensional space, and let $W_g \in \R^{d \times d_g}$ be trainable parameters that embed the target embedding $u_g$ in the same space. For every sub-window index $i,j \in \{1,...,n_v\}$, the visual-target attention potential $\phi^t_{g} (\cdot)$ at time $t$ takes the form:
\[
\phi^t_{g}(i,j) = \left \langle \frac{W_v v^t_{i,j}}{ \| W_v v^t_{i,j} \| }, \frac{W_g u_g}{\| W_g u_g \|} \right \rangle. 
\]
We apply $\ell_2$-normalization before the multiplication, i.e., we use the {\it cosine} similarity to compute the potential interaction between the target $u_g$ and the image sub-window~$v^t_{i,j}$. The corresponding attention probability distribution is attained by applying the softmax operation:  
\[
p^t_{g}(i,j) =\frac{ e^{\phi^t_g(i,j)} }{\sum_{s,t=1}^{n_v} e^{\phi^t_g(s,t)}}.
\]

See Figure~\ref{fig:architecture} for an example of the target attention probability distribution, $p_g^t(\cdot)$, for a target word ``toaster".
One can verify that the inferred  probability distribution is focused on the area in the image that contains the toaster.

\subsubsection{Action attention unit} 
\label{sec:action}
This unit gets as input the image and the last step's action distribution. In practice, the action probability distribution correlates with the agent's movement. 
The actor's actions are sampled from the policy $\pi_\theta(\cdot | s)$ that chooses an actions, given an egocentric RGB image $s \in S$. At time $t=1,...,k$, the agent samples an action $a_t$ from one of the six actions $a_t \in \{$\texttt{MoveAhead}, \texttt{RotateLeft},\texttt{RotateRight}, \texttt{LookDown}, \texttt{LookUp}, \texttt{Done}$\}$. At step $t$, the policy $\pi_\theta(\cdot | s)$ utilizes the actions distribution at time $t-1$, which we denote by $u^{(t-1)}_a \in \R^{d_a}$.

Similarly to the target attention unit, each of the $n_v \times n_v$ sub-windows of the image that is seen by the agent at time $t$ is encoded by a vector of length $d$, using the matrix~$W_v$. We also embed $u^{(t-1)}_a$ to the $d$ dimensional space by the learned matrix $W_a \in \R^{d \times d_a}$. For every sub-window index $i,j \in \{1,...,n_v\}$, the visual-action attention potential $\phi^t_a (\cdot)$ at time $t$ takes the form:
\[
\phi^t_{a}(i,j) = \left \langle \frac{W_v v^t_{i,j}}{ \| W_v v^t_{i,j} \| }, \frac{W_a u^{(t-1)}_a}{\| W_a u^{(t-1)}_a \|} \right \rangle. 
\]
Importantly, the action potential function considers the observed image at time $t$ and the preceding action (at time $t-1$).  
The corresponding attention probability distribution is attained by applying the softmax operation:  
\[
p^t_{a}(i,j) =\frac{ e^{\phi^t_{a}(i,j)} }{\sum_{s,t=1}^{n_v} e^{\phi^t_{a}(s,t)}}.
\]
An example of the action attention probability distribution $p_a^t(\cdot)$  appears in Figure \ref{fig:architecture}. 

\subsubsection{Memory attention unit} 
\label{sec:memory}
The memory attention unit summarizes the agent's experience and aims to focus on sections of the image based on the information already gathered in the episode.
For example, the agent should avoid focusing its attention on irrelevant areas that were previously explored. 
This unit gets as input the image and the agent's gathered experience in the episode up to the $t^{th}$ step, i.e., the  actions, the observed images, and the internal state representations that appear in the $(t-1)^{th}$ prefix of the agent's trajectory $(a_1,...,a_{t-1})$. This experience is represented by the hidden state of an LSTM-cell, whose input is the spatially attended observed image; its output is fed into the actor-critic module.  

As before, this unit learns a probability distribution function over the observed image at time $t$. The $n_v \times n_v$ sub-windows are encoded by vectors of length $d$ using the matrix $W_v$. The memory is extracted from the hidden state of an LSTM-cell at time $t-1$. We denote this state by $u^{(t-1)}_m \in \R^{d_m}$ and embed it in the $d^{th}$ dimensional space by the learned matrix $W_m \in \R^{d \times d_m}$. For every sub-window index $i,j \in \{1,...,n_v\}$, the visual-memory potential $\phi^t_{m} (\cdot)$ at time $t$ takes the form:
\[
\phi^t_{m}(i,j) = \left \langle \frac{W_v v^t_{i,j}}{ \| W_v v^t_{i,j} \| }, \frac{W_m u^{(t-1)}_m}{\| W_m u^{(t-1)}_m\|} \right \rangle. 
\]
The corresponding attention probability distribution is:  
\[
p^t_{m}(i,j) =\frac{ e^{\phi^t_{m}(i,j)} }{\sum_{s,t=1}^{n_v} e^{\phi^t_{m}(s,t)}}.
\]
The role of the memory attention probability distribution is demonstrated in Figure \ref{fig:architecture}. 

\subsubsection{Fused attention unit} 
\label{sec:fused}

The different attention units are constructed to capture different behaviors of the agent. However, these attention maps should be fused to summarize the target, the action and the memory attentions, which are represented by the respective probability distributions over the image. 

Naively, we can fuse these three probability distributions by normalizing their product $p^t_{g}(i,j) \cdot p^t_{a}(i,j) \cdot p^t_{m}(i,j)$. This allows to fuse the attention while accounting for each probability in a symmetric manner. However, this does not allow to learn the importance of each probability at time step $t$. 
Instead, we learn the importance of each probability at time $t$ by considering the hidden state of the LSTM-cell at time $t-1$, denoted by~$u_m^{(t-1)}$. 
Specifically, we learn the real-valued weight functions $\beta_g(u_m^{(t-1)}), \beta_a(u_m^{(t-1)}), \beta_m(u_m^{(t-1)})$ to fuse the different attention probability distributions at time $t$. We use the short hand notation $\beta_g, \beta_a, \beta_m$ for these functions and attain the fused attention probability distribution: 
\[
p^t(i,j) \propto \Big( p^t_{g}(i,j)^{\beta_g} p^t_{a}(i,j)^{\beta_a} p^t_{m}(i,j)^{\beta_m} \Big).
\label{eq:fuse}
\]

The fused attention is able to combine all attention probability distribution to a coherent distribution, see Figure~\ref{fig:architecture}. 


\begin{figure*}[t]
  \centering
  \large
    \begin{tabular}{cccc}
    \includegraphics[height=3.2cm]{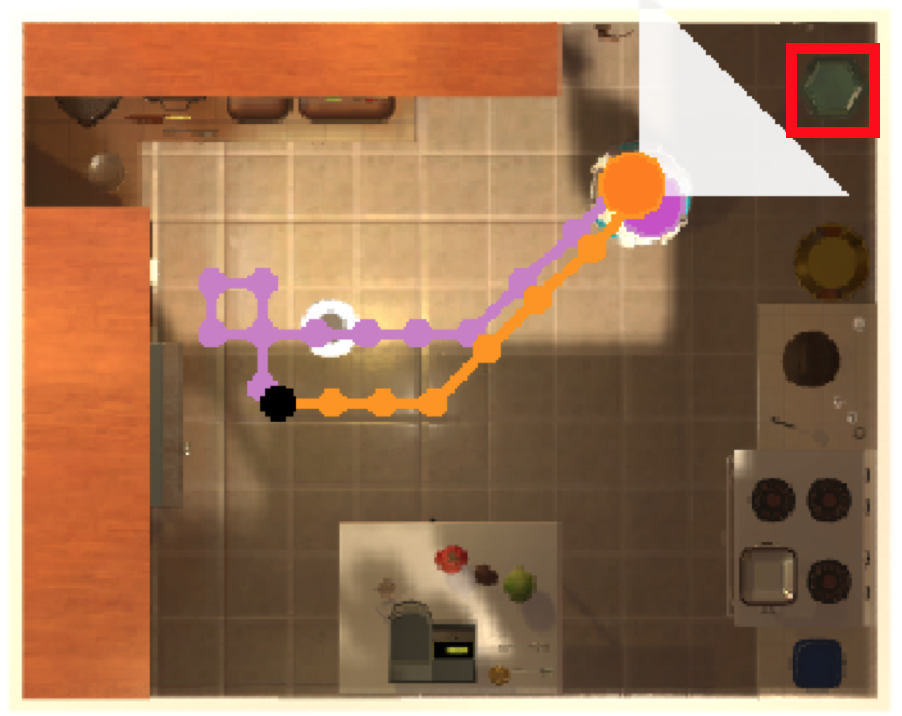} &
    \includegraphics[height=3.2cm]{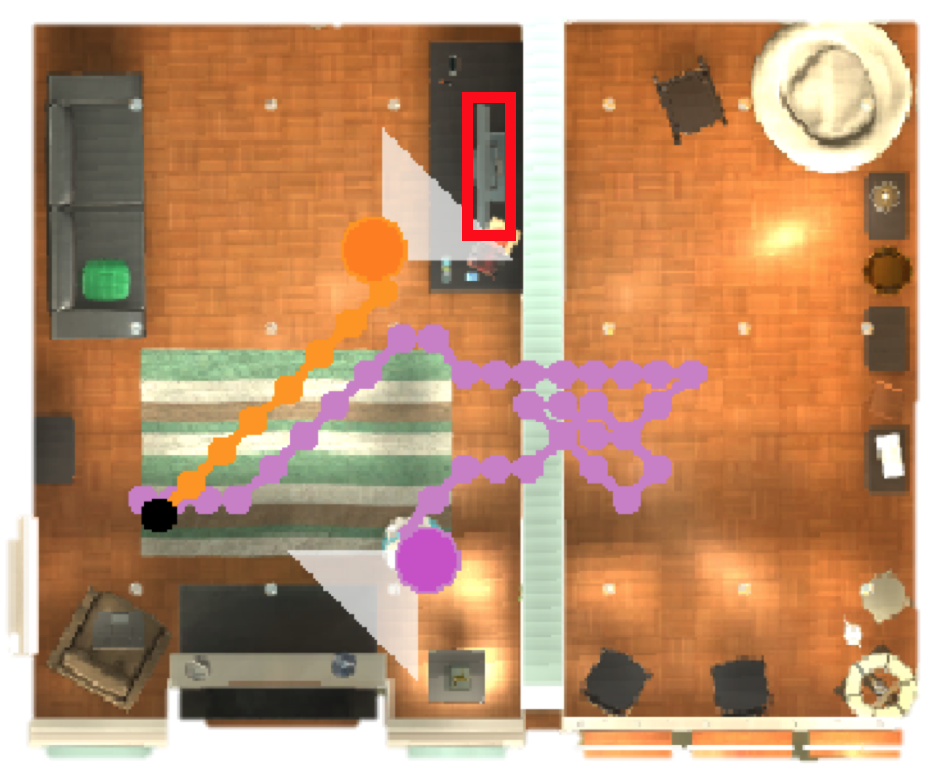} &
    \includegraphics[height=3.2cm]{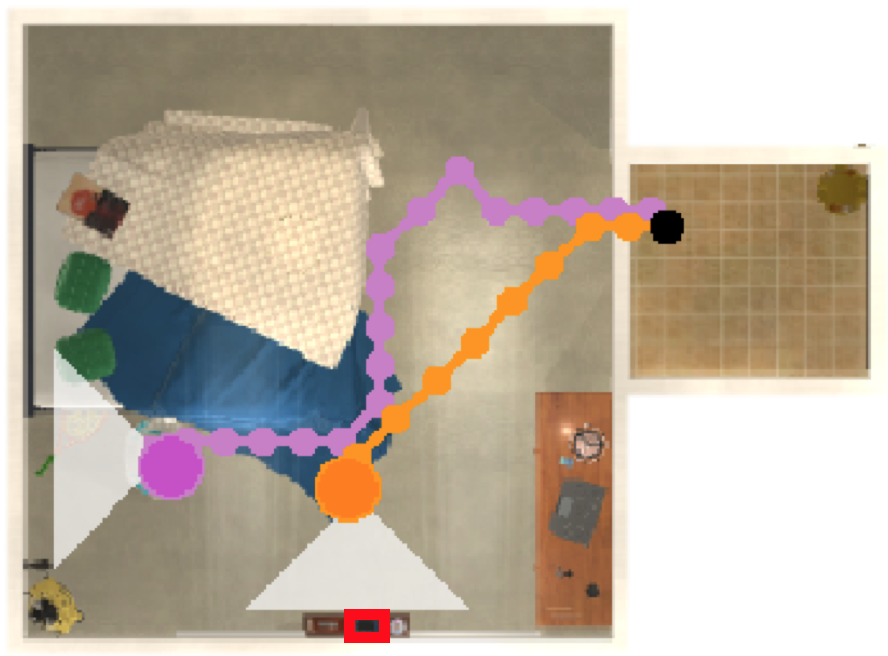} &
    \includegraphics[height=3.2cm]{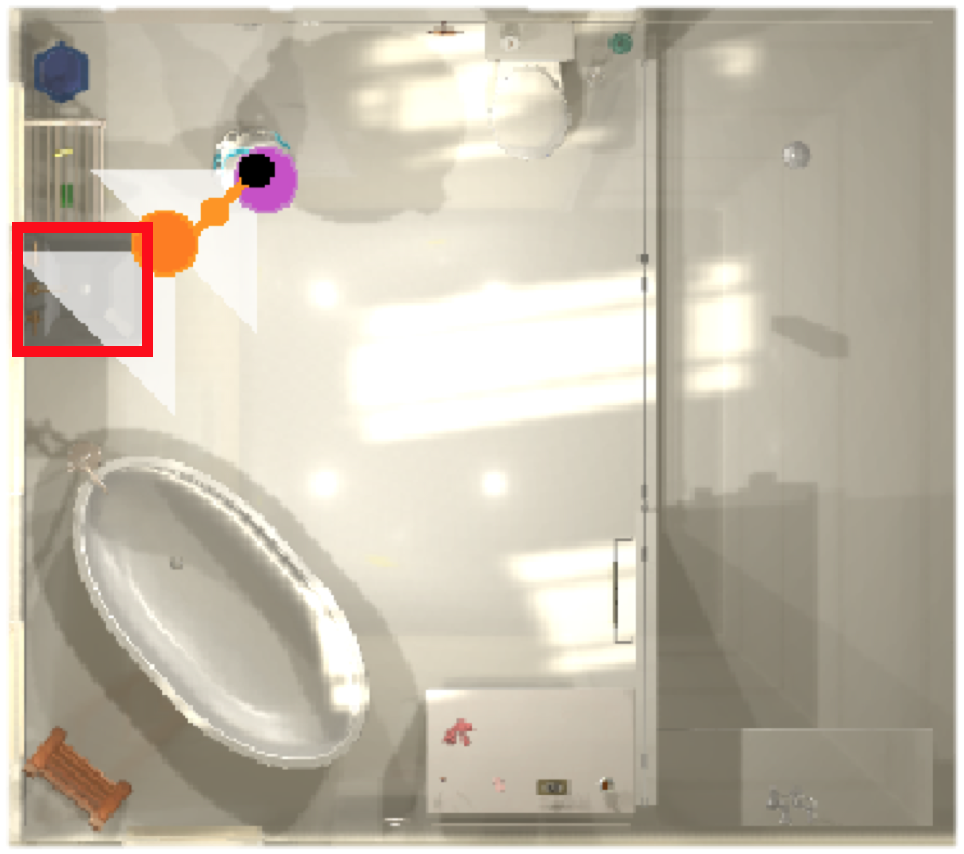} \\
   \small{(a) Both reached the goal;} & \small{(b) Reaching the goal thanks to} &
    \small{(c) Reaching the goal despite}  & \small{(d) Reaching the goal while}\\[-0.05in]  
    \small{our path is shorter} & \small{semantic external clues} & \small{target's irregular position} & \small{SAVN is too far from target}\\
  \end{tabular}
  \caption{
  {\bf Qualitative results.} 
  This figure compares our agent's trajectories to those of (SAVN)~\cite{Wortsman_2019_CVPR}.
   The starting position of the agent is drawn as a black circle and the target as a red box;
   the path of our agent is in orange and SAVN's path is in magenta; 
   accordingly the large orange/magenta points show the final locations of the respective agents.
(a) While both agents reach the target (a garbage can), our path is shorter, since our attention model enables our agent to gather information on the scene early on. 
This is also expressed in the SPL evaluation in Table~\ref{table:results}.
(b) Our agent found the TV, whereas SAVN misses it, probably due to ignoring the spatial cues in the living room, such as the carpet or the TV stand; 
(c) The alarm clock is situated in irregular position (near the dresser) and is very small.
While our agent is able to focus on a small region in the observation and locate the alarm clock, SAVN's agent continues its search near the bed and misses it.
The lack of attention unit in SAVN results in more emphasis on object locations than on visual characteristics.
(d) Our agent found a sink, whereas SAVN’s agent stopped too far, and yet declared it found the target; i.e.  the distance estimation is wrong.  
    }
  \label{fig:qaulitative}
\end{figure*}

\section{Experimental Validation}
\label{sec:experiments}
We follow Wortsman et al.~\cite{Wortsman_2019_CVPR} and train and evaluate our models using the AI2-THOR~\cite{zhu2017icra} environment with their scenes from the four room categories: kitchen, living room, bedroom and bathroom. For each room type, we use the same $20/5/5$ split of train/validation/test for a total of $120$ scenes. The objects in their scenes, per room type are: 1)~Living room: pillow, laptop, television, garbage can, box, and bowl. 2) Kitchen: toaster, microwave, refrigerator, coffee maker, garbage can, box, and bowl. 3) Bedroom: plant, lamp, book, and alarm clock. 4) Bathroom: sink, toilet paper, soap bottle, and light switch. We also use the reward function of~\cite{Wortsman_2019_CVPR}, with reward of $5$ for finding the object and $-0.01$ for taking a step. We learn a policy for this reward using actor-critic reinforcement learner with an advantage function and $12$ synchronous agents (A2C). 
The scene, initial state of the agent and the target object were chosen by~\cite{Wortsman_2019_CVPR} and for each training run we select the model that performs best on the validation set in terms of success.

The agent moves with the  \texttt{MoveAhead} action. The \texttt{RotateLeft} and \texttt{RotateRight} actions occur in increments of $45^\circ$, while the  \texttt{LookDown} and \texttt{LookUp} actions tilt the camera by $30^\circ$. During training, the maximal trajectory consists of $30$ actions and during validation and testing the maximal trajectory is limited to $200$ actions to living rooms and $100$ actions to other room types. 
The agent successfully completes a navigation task if it performs the \texttt{Done} action when an instance from the target object class is within $1$ meter from the agent's camera and within the agent's field of view. 

The methods are evaluated using both Success Rate and Success weighted by Path Length (SPL). Success is defined as $\frac{1}{N}\sum_{i=1}^N \mathcal{S}_i$  where $N=1000$ is the number of episodes ($250$ episodes for each scene type in the test set) and $\mathcal{S}_i$ is a binary indicator of success in episode $i$. The SPL is defined as $\frac{1}{N}\sum_{i=1}^N \mathcal{S}_i \frac{L_i}{\max(P_i, L_i)}$ and it measures the quality of the agent's path when it succeeds in finding the object in episode $i$, where $P_i$ denotes path length and $L_i$ is the length of the optimal trajectory to any instance of the target object class in that scene. As the behavior of the agent's policy is different for short and long paths. we also refer to trajectories where the optimal path length is at least $5$ and denote this by $L \geq 5$ ($L$ refers to optimal trajectory length).

\begin{table}[t]
\centering
  \begin{tabular}{ |l||c|c|c|c|  }
      \hline
           Architecture  & SPL & Success & SPL  & Success \\
             &  &  &  $L\ge 5$ &  $L \ge 5$\\
       \hline
         Scene Prior \cite{yang2018visual} &15.47 &35.13 &11.37 &22.25 \\
         SAVN \cite{Wortsman_2019_CVPR} & 16.15 &40.86 &13.91 &28.70   \\
         Ours (A3C) &16.99 &43.20 &15.51 &31.71 \\
         {\bf Ours (A2C)} &{\bf 17.88} & {\bf 46.20} &{\bf 15.94} & {\bf 32.63} \\
      \hline
  \end{tabular}
\caption{{\bf Quantitative results.} Our best results are attained for synchronous actor-critic learner (A2C).
However, our asynchronous learner (A3C) outperforms the asynchronous learner of~\cite{Wortsman_2019_CVPR} as well. }
\label{table:results}
\end{table}

Table \ref{table:results} compares our results to the state-of-the-art and shows improvement over previous works in terms of both success rate and path length (SPL), for short paths as well as for long paths. During our experimental validation we noticed that synchronization is important to get stable results over different platforms and GPUs. Empirically, our best results are attained for synchronous actor-critic learner (A2C) rather than asynchronous actor-critic learner (A3C). Nevertheless, our asynchronous learner outperforms asynchronous learner of SAVN \cite{Wortsman_2019_CVPR}.

Figure \ref{fig:qaulitative} shows different scenarios and compares the behavior of our agent to that of SAVN.
It demonstrates how the added spatial attention information allows our agent to better navigate in the $3D$ Euclidean space. 
In particular, in all these scenarios our agent reaches the goal efficiently whereas SAVN's agent
(a)~takes longer to find the target object;
(b) misses the target, as its non-optimal trajectory sets it on a path for which the object is not in view;
(c)~misses the target due to being small and situated in a irregular position;
(d) stops too far.
Our attention model enhances the agent's ability to notice and use visual clues. 
This is turn, manifests in better object mapping and path planing.

\begin{figure}[t]
\centering
\includegraphics[width=0.48\textwidth]{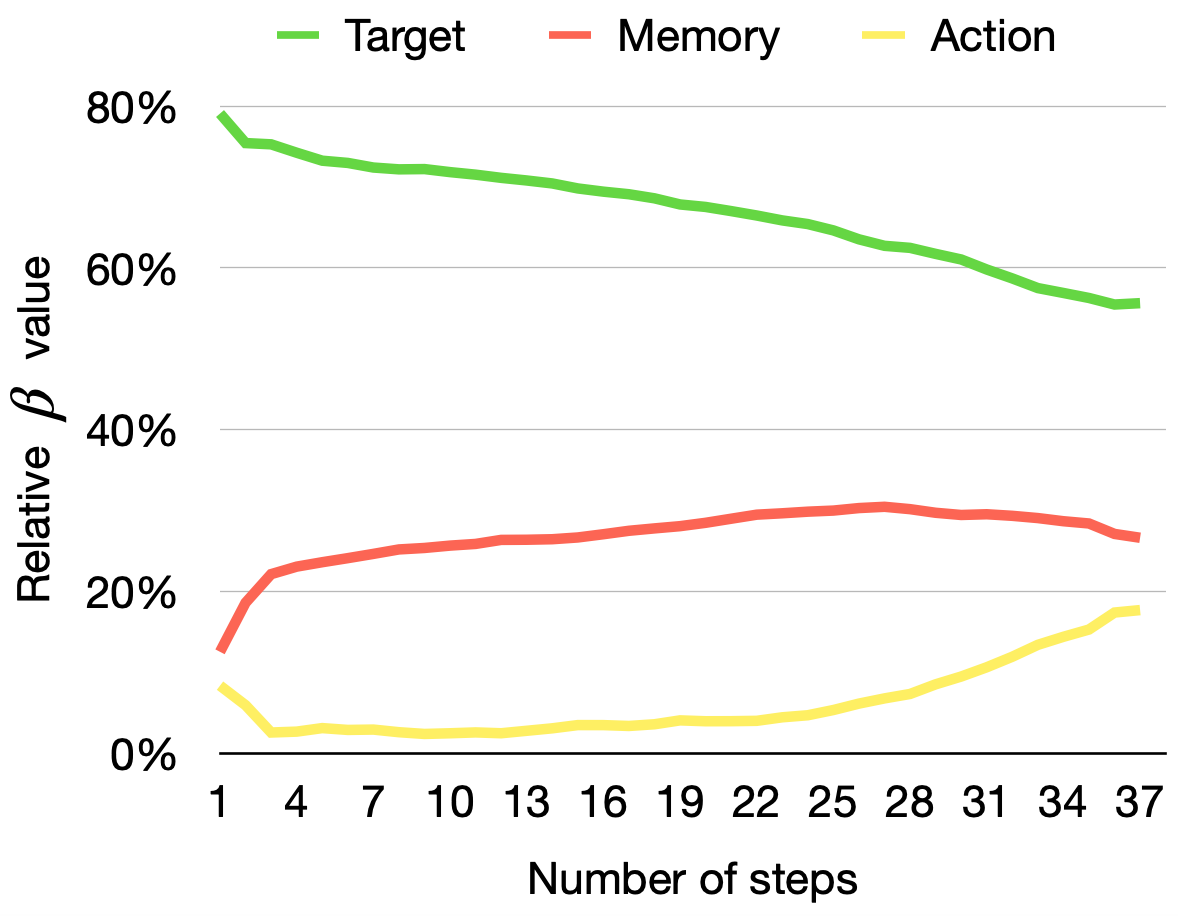}
\caption{{\bf The $\beta$ weights.}
This graph shows how $\beta_g(u_m^t), \beta_a(u_m^t), \beta_m(u_m^t)$ of the target/action/memory attention units change along the navigation. The $y$-axis is proportion of the respective unit, e.g., $|\beta_g(u_m^t)| /  \sum_{i \in g,a,m} |\beta_i(u_m^t)|$ and the $t^{th}$ tick of the $x$-axis is the average of the respective proportion for all test episodes in their $t^{th}$ step. We capped $t$ by $37$ as there are negligible number of trajectories that have more than $37$ steps.
}
\label{fig:beta}
\end{figure}

Recall that our spatial attention is based on three attention probability distributions that are based on the target, the agent's previous action and the agent's memory, which is represented in its LSTM-cell. 
These three distributions are fused to a single attention probability model,  which weighs the three factors according to the agent's LSTM-cell. 
Figure~\ref{fig:beta} provides a quantitative assessment of the weights, $\beta_g(u_m^t), \beta_a(u_m^t), \beta_m(u_m^t)$, which control the fused distribution in Equation \ref{eq:fuse}. 
We also compared the aforementioned fused module with a more expressive baseline that concatenates the three attended feature maps and use it as the hidden feature for the actor-critic module. While our system requires less parameters (7.5M vs. 20.4M) it performs better than the expressive baseline: our SPL is $31\%$ better, Success is $22\%$ better, SPL $L \ge 5$ is $41\%$ better, and Success $L \ge 5$  is $30\%$ better than the baseline.


 We also tested our attention module in the densely annotated setting of \cite{du2020learning}. This setting use the simulator information to extract detected objects which allows us to integrate their information to the attention module. In this setting we our attention module, built over the architecture of \cite{du2020learning}, achieve an improvement of $9.2\%$ in Success rate, $9.1\%$ in SPL, $13.7\%$ in Success $L \ge 5$, $13.2\%$ in SPL $L \ge 5$, when measured over the $4000$ test scenes. Also, since our attention adds spatial information to our agent, we note that the number of test images in which we detected the target object is $15.8 \%$ higher in our case than when using \cite{du2020learning}.

\paragraph{Ablation study.}
The aim of the ablation study is to verify the validity of our attention unit and the importance of its different components. 
Table \ref{table:ablation} compares our attended embedding with the state-of-the-art multi-head embedding of the transformer~\cite{vaswani2017attention} (MHA in the table)).
The difference in performance is related to the different embedding strategies of the two methods. The transformer embeds the data using learned probability distributions (of key and query) and learns representation (value).
While this embedding is very effective in language processing, its embedding ignores spatial information in visual tasks. 

\begin{table}[t]
\centering
  \begin{tabular}{ |p{2cm}||p{1cm}|p{1cm}|p{1cm}|p{1cm}|  }
      \hline
           Architecture  & SPL & Success & SPL $L\ge 5$ & Success $L \ge 5$\\
       \hline
         SAVN \cite{Wortsman_2019_CVPR} & 16.15 &40.86 &13.91 &28.7   \\
         Ours (MHA) &9.80 &30.70 &8.29 &20.36   \\
         Ours w/o $p_g$ & 9.41 & 29.60 &7.93 & 19.01 \\
         Ours w/o $p_a$ &14.41 &41.00 &13.07 &29.64 \\
         Ours w/o $p_m$ &15.39 &45.6 &14.07 &{\bf 33.08} \\
         Ours $\beta=1$ &13.88 & 38.80 &11.30 & 26.35 \\
         {\bf Ours (A2C)} &{\bf 17.88} & {\bf 46.20} &{\bf 15.94} & 32.63\\
      \hline
  \end{tabular}
\caption{{\bf Ablation study.} The table presents the significance of our attention module and its separate units. Replacing our attention unit with multi-head attention of \cite{vaswani2017attention} (MHA) decreases our success rate by $8\%$ for all paths and $7\%$ for long paths ($L \ge 5)$. Omitting the target attention unit decreases our success rate by about half, while the contributions of the action and memory units are comparable. Fixing the balance $\beta$ of the various attention units during a test episode decreases our success rate by more than $20\%$.}
\label{table:ablation}
\end{table}

Table \ref{table:ablation} also shows how our fused attention model performs when we take out its target/action/memory attention components. 
One  can verify that  each of the components is vital to gain good performance. The target attention unit is the most important module, as we are focused on target-driven visual navigation, while the contribution of the action \& memory units is comparable. 
We also see the importance of changing the balance $\beta$ of the various attention units, as when we use a fixed $\beta=1$ during learning and testing, the results deteriorate.

\begin{figure}[t]
\centering
    \begin{tabular}{cc}    
    \includegraphics[width=0.2\textwidth]{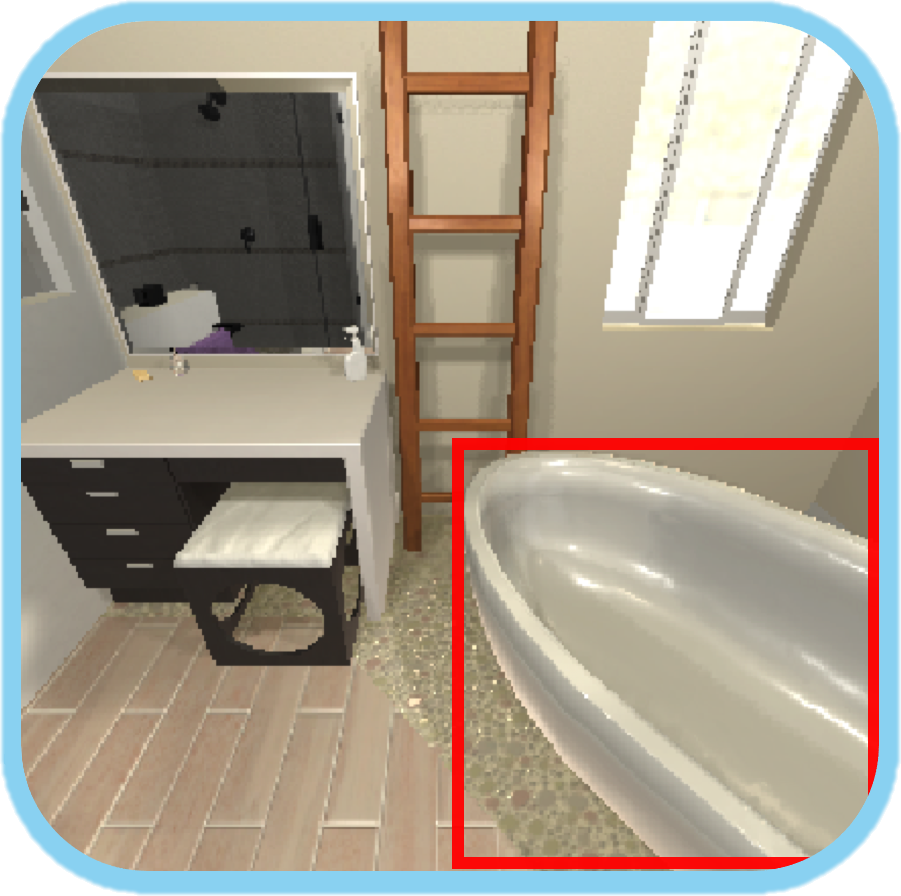} & 
    \includegraphics[width =0.2\textwidth]{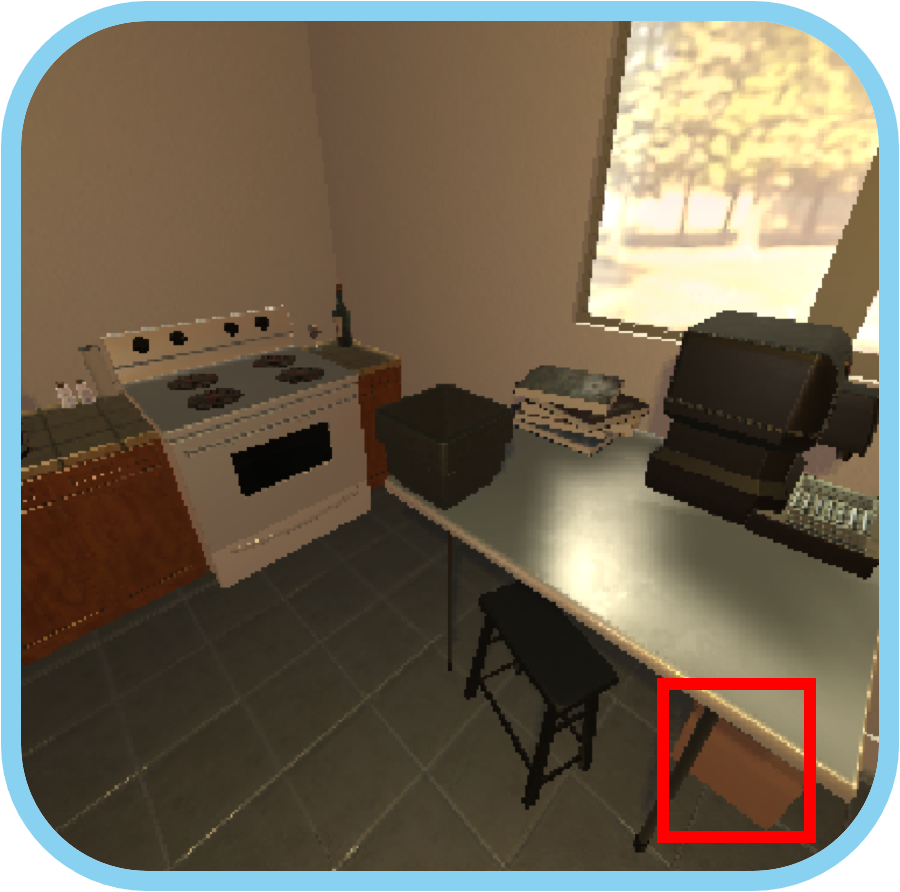} 
    \\
    (a) erronous classification&
    (b) hidden object
  \end{tabular}
\caption{{\bf Limitations.}
(a) The bathtub is classified as a sink, due to shape similarity.
(b)~The agent failed to find the partially-hidden box under the table.
}
\label{fig:limit}
\end{figure}

\paragraph{Limitations.}
Figure~\ref{fig:limit} shows two cases where our system fails.
In~(a), the agent is looking for a sink and thinks it found it despite of the fact that it actually found a bathtub. 
We note that this specific bathtub is similar in shape to some sinks in the training data.
In~(b), the agent is looking for a box, which is partially hidden.
The visible part of the box is insufficient for our agent to conclude it found the object. 

\section{Conclusion}
This work presented an end-to-end reinforcement learning for visual navigation. 
Our framework is based on a novel attention probability model that suits visual navigation, as it encodes both semantic information about the observed objects and spatial information about their place.   
Specifically, the attention model consists of three components: target, action and memory.
The framework is shown to achieve SOTA results on commonly-used scenarios.

Our results are achieved using only RGB images.
In the future, RGBD images can be utilized, as suggested by~\cite{chaplot2020object}.
This has the potential to shorten the path, as the distance requirement can be achieved more precisely and obstacles may be more easily avoided.

\newpage
\bibliographystyle{ieee_fullname}
\small
\bibliography{egbib}

\end{document}